\documentclass[a4paper,num-refs,gigabyte]{oup-contemporary}

\journal{gigabyte}

\usepackage{graphicx}
\usepackage{amsmath}
\usepackage{amsfonts}
\usepackage{wrapfig}
\usepackage{needspace}
\usepackage{multirow}
\usepackage{siunitx}
\usepackage{eso-pic}
\usepackage{tcolorbox}
\usepackage{tabularx}
\usepackage{enumitem}
\usepackage{indentfirst}

\makeatletter
\newcommand{\finishwrapfigure}{%
  \par
  \ifnum\c@WF@wrappedlines>\@ne
    \vspace*{\dimexpr\numexpr\c@WF@wrappedlines-1\relax\baselineskip\relax}%
  \fi
  \WFclear
  \addvspace{\intextsep}%
}
\makeatother

\definecolor{paperbackground}{HTML}{FBFCFC}
\AtBeginDocument{\pagecolor{paperbackground}}

\newcommand{\themebf}[1]{\textcolor{jcolour!80!black}{#1}}

\RenewEnviron{abstract}{%
  \begin{tcolorbox}[
    colback=jcolour!9,
    colframe=jcolour!28,
    boxrule=0.35pt,
    arc=6pt,
    outer arc=6pt,
    boxsep=0pt,
    left=11pt,
    right=11pt,
    top=7pt,
    bottom=7pt,
    before skip=0pt,
    after skip=4pt
  ]
    \fontsize{7.8pt}{10pt}\selectfont
    \setlength{\parindent}{12pt}%
    \setlength{\parskip}{0pt}%
    \noindent\textbf{\color{jcolour}Abstract.}\hspace{0.55em}%
    \BODY\par
  \end{tcolorbox}%
}

\ifPDFTeX
  \newcommand{\TencentTitleFont}{\bfseries}
\else
  \usepackage{fontspec}
  \IfFileExists{TencentSansW7.ttf}{%
    \setmainfont[Path=./]{TencentSansW7.ttf}%
    \setsansfont[Path=./]{TencentSansW7.ttf}%
    \newfontfamily{\TencentTitleFont}[Path=./]{TencentSansW7.ttf}%
  }{%
    \newcommand{\TencentTitleFont}{\bfseries}%
  }%
\fi

\newcommand{\wepmpagenumber}{%
  \fontsize{7.8pt}{10pt}\selectfont%
  \color{jcolour}\textbf{\thepage}%
}
\fancypagestyle{plain}{%
  \fancyhf{}%
}
\AddToShipoutPictureFG{%
  \AtPageLowerLeft{%
    \raisebox{15pt}[0pt][0pt]{%
      \makebox[\dimexpr\paperwidth-39pt\relax][r]{\wepmpagenumber}%
    }%
  }%
}

\renewcommand{\thesection}{\arabic{section}.}
\renewcommand{\thesubsection}{\arabic{section}.\arabic{subsection}}
\renewcommand{\thesubsubsection}{\thesubsection.\arabic{subsubsection}}
\titleformat{\section}{\fontsize{10pt}{12pt}\bfseries\color{jcolour}}{\thesection}{0.6em}{}
\titleformat{\subsection}{\fontsize{8.5pt}{11pt}\bfseries\color{jcolour}}{\thesubsection}{0.6em}{}
\titleformat{\subsubsection}{\fontsize{8pt}{10pt}\bfseries\color{jcolour}}{\thesubsubsection}{0.6em}{}

\title{{\fontsize{13pt}{17pt}\selectfont\TencentTitleFont WePM: Hierarchical Conflict Reconciliation for Unified MLLMs}\\[7pt]
{\color{jcolour!55!black}\fontsize{10pt}{14pt}\selectfont Unified Pocket Models for In-App Multimodal Intelligence}}

\newcommand{\wechatteamauthor}{%
  {\raggedright
  \raisebox{-0.42\height}{\includegraphics[height=26pt]{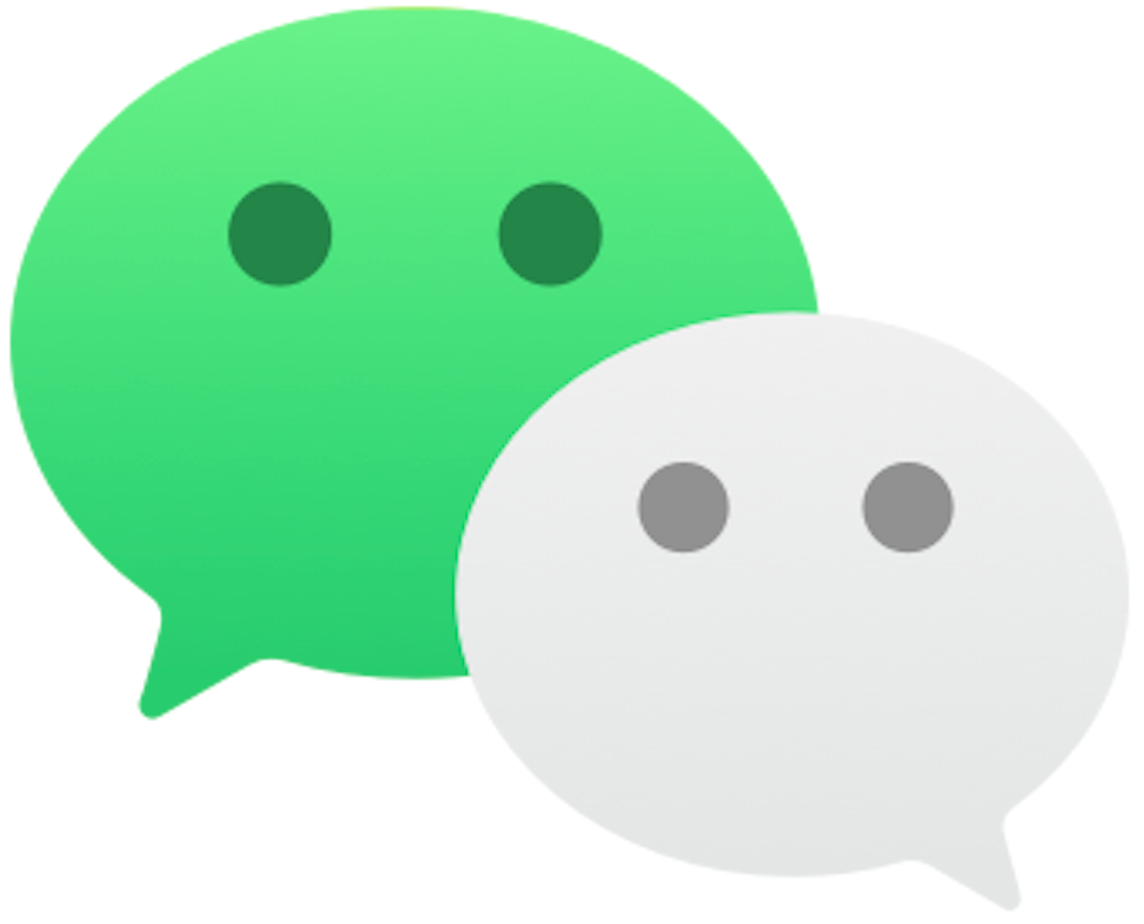}}%
  \hspace{0.7em}%
  {\fontsize{13pt}{17pt}\selectfont\TencentTitleFont WeChat AI}%
  \par}%
}

\newcommand{\wepmhuggingfaceurl}{https://huggingface.co/collections/tencent/wepocketmodel}
\newcommand{\wepmgithuburl}{https://github.com/Tencent/WePocketModel}
\newcommand{\wepmauthorblock}{%
  \begin{center}
    {\fontsize{9.2pt}{12pt}\selectfont
      Zhiqiang Yuan$^{1,\clubsuit,\dagger}$, Qiwei Yan$^{2,\clubsuit}$, Liwei Yang$^{3,\clubsuit}$,
      Jinzhe Liu$^{2}$, Yang Liu$^{1}$, Zhongyan Hou$^{2}$\\[2pt]
      Liyan Xu$^{1}$, Peixiang Luo$^{1}$, Jiapei Zhang$^{1}$, Ying Deng$^{1}$,
      Jie Zhou$^{1}$, Jinchao Zhang$^{1,\heartsuit}$\par}
    \vspace{10pt}
    {\fontsize{7.8pt}{10pt}\selectfont
      $^{1}$WeChat AI, Tencent \quad
      $^{2}$University of Chinese Academy of Sciences \quad
      $^{3}$Xi'an Jiaotong University\par}
    \vspace{3pt}
    {\fontsize{7.4pt}{9pt}\selectfont
      $\clubsuit$ Equal contribution. \quad $\dagger$ Tech Lead. \quad $\heartsuit$ Program Lead.\par}
    \vspace{5pt}
    {\fontsize{8pt}{10pt}\selectfont
      \href{\wepmgithuburl}{\raisebox{-0.22\height}{\includegraphics[height=10.5pt]{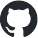}}\hspace{0.35em}{\color{jcolour!80!black}GitHub}}\quad
      \href{\wepmhuggingfaceurl}{\raisebox{-0.28\height}{\includegraphics[height=11pt]{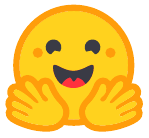}}\hspace{0.35em}{\color{jcolour!80!black}Hugging Face}}\par}
  \end{center}%
}

\makeatletter
\AtBeginDocument{%
\renewcommand{\@maketitle}{%
  \vspace*{0.5pc}%
  \noindent\wechatteamauthor
  \vspace{6pt}%
  \noindent{\color{jcolour!35}\rule{\textwidth}{0.7pt}}\par
  \vspace{12pt}%
  \begin{center}
    {\color{jcolour}\@title\par}%
  \end{center}%
  \vspace{7pt}%
  \wepmauthorblock
  \vspace{5pt}%
}%
}
\makeatother

\RenewEnviron{frontmatter}{\BODY}

\AddToShipoutPictureBG{%
  \AtPageCenter{%
    \makebox(0,0){%
      \rotatebox{35}{%
        {\fontsize{82pt}{96pt}\selectfont\TencentTitleFont\color{jcolour!12}Wechat AI}%
      }%
    }%
  }%
}

\papercat{Paper}

\runningauthor{Yuan et al.}

\jvolume{00}
\jnumber{0}
\jyear{2020}

\begin{document}

\begin{frontmatter}
\maketitle
\begin{abstract}
Bringing multimodal intelligence from the cloud to users’ devices keeps sensitive data local and substantially reduces cloud inference costs, but deploying a specialized model for every capability duplicates vision--language capacity under tight memory and compute budgets. In this paper, we first introduce and formalize the concept of heterogeneous capability density (HCD), which characterizes how broadly and effectively a model serves heterogeneous task families per unit of resident inference memory. Achieving high HCD across representation, generation, and decision tasks presents two fundamental challenges. First, these capabilities require visual evidence at different spatial granularities, yet must operate within a fixed token budget. Second, learning from separately collected task corpora requires coordinating potentially conflicting updates to shared parameters. Constructing cross-task paired supervision to support such coordination is costly and often ambiguous for open-ended inputs; directly combining task gradients, however, can lead to interference and imbalanced task contributions due to differences in gradient scale, loss weighting, and exposure frequency.
To address these challenges, we present WePM, a family of pocket models that unify multimodal representation, generation, and decision within a single vision--language backbone. Multi-Granularity Visual Budgeting (MGVB) allocates a fixed visual-token budget across global, regional, and fine-grained banks to reconcile heterogeneous evidence requirements.Unified Listwise Knowledge Distillation (ULKD) enriches separately collected representation and generation data with soft teacher guidance, avoiding costly cross-task paired annotations and arbitrary hard targets for open-ended inputs. Stable and Efficient Multitask Optimization (SEMO) regulates each task’s contribution to shared-parameter updates through explicit loss weights and exposure schedules, while sequential gradient accumulation executes the joint update within training-memory constraints. Built on these mechanisms, WePM uses task-appropriate paths to produce normalized embeddings, autoregressively decoded responses, and schema-constrained decisions.
For a direct comparison with existing unified methods, we first evaluate WePM in the shared joint representation--generation setting, where it outperforms all compared baselines across five evaluation groups and raises the Joint score from the strongest baseline’s 38.15 to 52.74. In single-task evaluations, WePM-4B achieves the highest average performance across all three capability families, scoring 72.86 over 30 embedding benchmarks, 62.41 over 39 generation benchmarks, and 61.11 over seven typed-decision benchmarks. These results demonstrate that WePM maintains strong and balanced task-specific performance while unifying heterogeneous capabilities. Compared with the separate-model stack with the highest HCD, WePM achieves slightly higher balanced performance with 61.9\% less resident weight memory, resulting in a 2.64-fold higher HCD. Together, these results show that diverse multimodal capabilities need not come at the cost of a large stack of specialized models, positioning WePM as a promising foundation for capable on-device AI under tight memory constraints.
\end{abstract}


\vspace{6pt}
\noindent\begin{minipage}{\textwidth}
  \centering
  \includegraphics[width=0.8\linewidth]{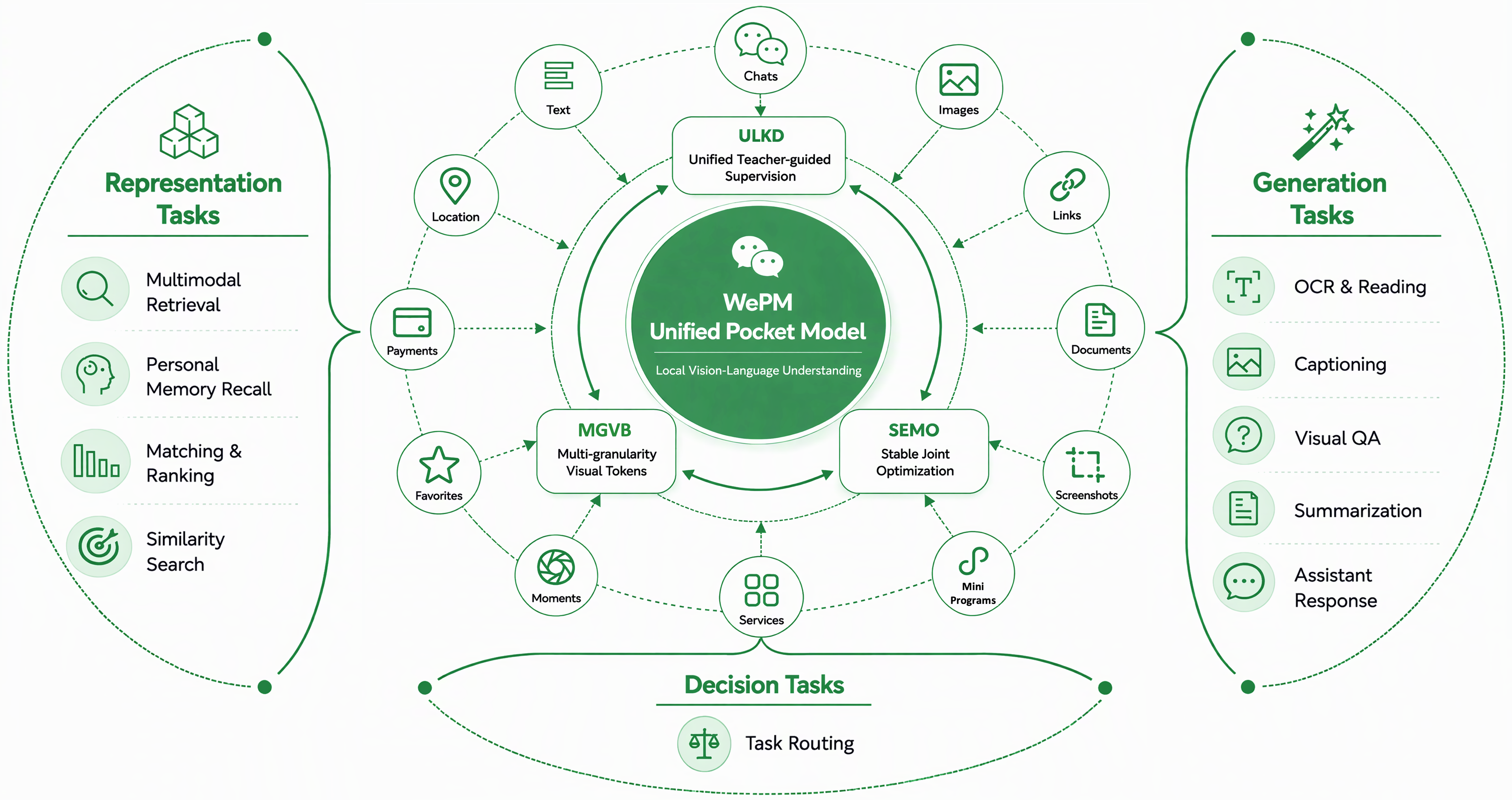}
  \captionof{figure}{Overview of WePM. WePM unifies heterogeneous local vision-language signals into a compact pocket model, supporting representation-oriented tasks such as multimodal retrieval and personal memory recall, generation-oriented tasks such as OCR, captioning, visual question answering, summarization, and assistant responses, and decision-oriented tasks such as schema-constrained option selection, ordinal scoring, and binary judgment.}
  \label{fig:wepm-overview}
\end{minipage}
\end{frontmatter}
\clearpage

\section{Introduction}
\label{sec:introduction}

One important direction for the next stage of AI is to bring intelligence from the cloud to users’ devices, helping users get things done in their daily lives while keeping sensitive data local and substantially reducing cloud inference costs. Such assistance spans heterogeneous capabilities, from finding and understanding personal content to producing useful responses and making context-dependent choices as interactions unfold. Yet, unlike cloud servers, on-device systems operate under tight constraints on system RAM, GPU memory, and compute. Deploying a separate specialized model for each capability duplicates vision--language capacity and inflates the resident resource footprint. This leads to a central question: how can a single compact model support more heterogeneous tasks, perform well on each of them, and remain within a limited on-device memory budget? In this paper, we investigate this question by unifying representation, generation, and decision within one on-device multimodal model.

To formalize this objective, we first introduce the concept of \themebf{\emph{Heterogeneous Capability Density}} (HCD), which characterizes the range and quality of capabilities that a model can expose per unit of resident inference memory. Let $K$ be the number of predefined task families that reach their respective performance thresholds, $M$ the total resident inference memory required to support these families, $p_i$ the normalized performance of family $i$, and $w_i$ its application-specific importance. We define
\[
\mathrm{HCD}=\frac{K}{M}\,H_w(p_1,\ldots,p_K),\qquad
H_w(p_1,\ldots,p_K)=\frac{\sum_{i=1}^{K}w_i}{\sum_{i=1}^{K}w_i/p_i},
\]
where $H_w$ is the weighted harmonic mean of the family-level scores. Task heterogeneity is defined at the capability-family level rather than the benchmark level: tasks that share the same fundamental output form and functional objective belong to the same family. For example, OCR and VQA are both treated as generation tasks and therefore do not separately increase $K$. The weights can reflect an application's demand for different capabilities; in this work, we assign equal weights to all families to encourage uniform coverage. HCD thus measures how broadly and effectively a model can serve an application within a fixed memory budget: a higher HCD indicates that the application can expose more capable functions without proportionally increasing its resident memory footprint.

Achieving high HCD across heterogeneous tasks---such as representation, generation, and decision---is far from straightforward. These capabilities differ not only in their output semantics but also in the visual evidence and training signals they require. Representation maps independently encoded inputs to normalized embeddings whose similarities must remain stable across indexed items and future queries. Generation instead requires hidden states to remain sensitive to the current instruction, fine-grained visual evidence, and an autoregressively growing answer prefix. Decision evaluates the current state against caller-specified options or criteria and produces a constrained choice, ordinal score, or binary judgment. Unifying these capabilities within one compact backbone therefore introduces two fundamental difficulties: they compete for visual evidence at different spatial granularities, and their task-specific objectives can interfere when updating the same parameters.

We organize these difficulties into a granularity conflict and a shared-optimization conflict. The \themebf{granularity conflict} concerns which evidence should be retained under a fixed visual-token budget: coarse semantics may suffice for broad retrieval, whereas hard-negative discrimination, detailed generation, and visually grounded decisions may require regional or fine-grained evidence. The optimization conflict arises because representation, generation, and decision data are typically collected in separate corpora and consumed through task-specific batches. An update that improves one objective can therefore degrade another, while unequal gradient scales and update frequencies can allow one task to dominate the shared parameters. Two natural responses expose further challenges. Constructing cross-task paired annotations leads to a \themebf{supervision conflict}, because such annotations are expensive and open-ended inputs may not admit a unique target in another output space. Integrating task gradients instead leads to a \themebf{magnitude conflict}, because each task's effective contribution depends jointly on its native gradient scale, loss coefficient, and exposure frequency.

To address these conflicts, we develop \themebf{WePM}, which coordinates shared capacity at the levels of visual evidence, task supervision, and optimization. \themebf{Multi-Granularity Visual Budgeting} addresses the granularity conflict by allocating a fixed visual-token budget across global, regional, and fine-grained banks, conditioning selection on the available text while retaining query-independent indexing. \themebf{Unified Listwise Knowledge Distillation} addresses the supervision conflict by transferring soft generation and representation preferences without requiring exhaustively paired annotations or imposing unique hard targets on open-ended inputs. \themebf{Stable and Efficient Multitask Optimization} addresses the magnitude conflict by explicitly controlling fixed loss coefficients and task exposure, while using sequential gradient accumulation to realize the specified joint update under constrained training memory.

\begin{figure*}[!b]
  \centering
  \includegraphics[width=0.95\textwidth]{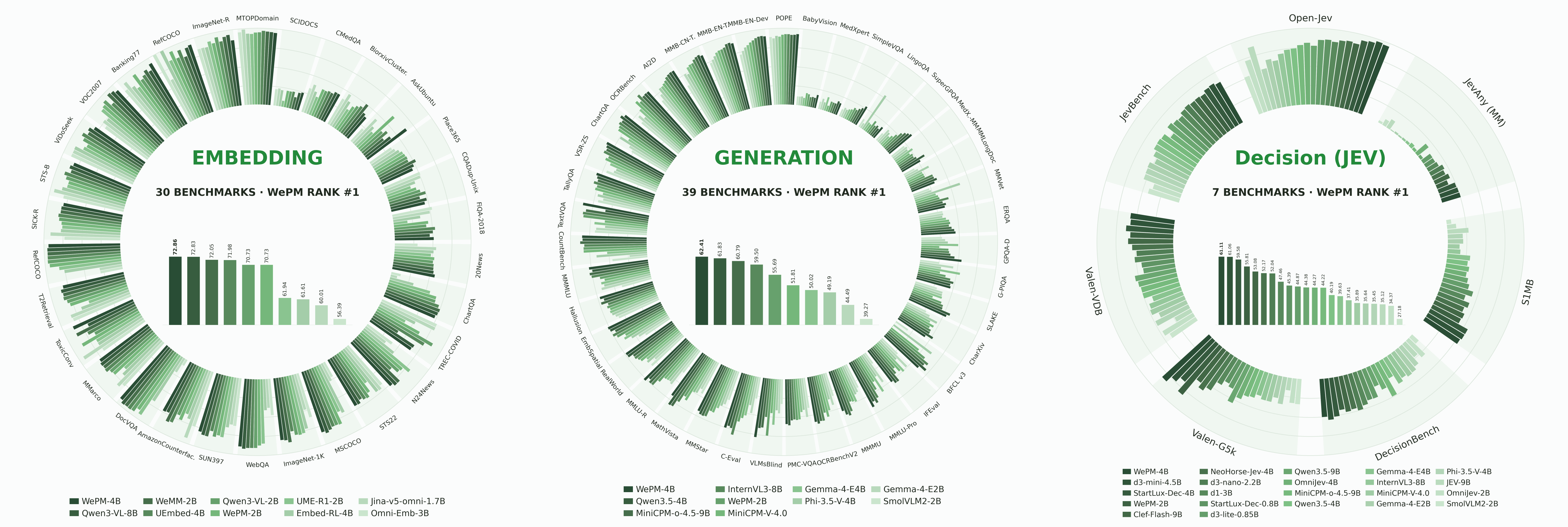}
  \caption{Single-task evaluation across the three capability families. The radial plots compare WePM with competing models on 30 embedding, 39 generation, and 7 decision benchmarks, while the inset bar charts summarize model-level average performance. WePM ranks first in all three capability families.}
  \label{fig:single-task-results}
\end{figure*}

Built on these mechanisms, WePM exposes three task-appropriate outputs from one vision--language backbone. Its representation path projects the final valid hidden state after multimodal prefill into a normalized embedding, while its generation path retains standard autoregressive decoding. Its decision path introduces a schema-conditioned decision slot and lightweight typed readouts that score caller-defined options or criteria in a single forward pass. By sharing the visual encoder and multimodal backbone, the three paths avoid the duplicated resident capacity of separate models while preserving the output semantics required by each task.

We instantiate WePM with a Qwen3.5~\cite{qwen3.5} backbone and evaluate its representation, generation, and typed-decision capabilities. For a direct comparison with existing unified methods, we first evaluate WePM in the shared joint representation--generation setting, where it outperforms all compared baselines across five evaluation groups and raises the Joint score from the strongest baseline's 38.15 to 52.74. In single-task evaluations, WePM-4B achieves the highest average performance across all three capability families, scoring 72.86 over 30 embedding benchmarks, 62.41 over 39 generation benchmarks, and 61.11 over seven typed-decision benchmarks. Compared with the separate-model stack with the highest HCD, WePM achieves slightly higher balanced performance with 61.9\% less resident weight memory, resulting in a 2.64-fold higher HCD (0.244 versus 0.092). Together, these results show that diverse multimodal capabilities need not come at the cost of a large stack of specialized models, positioning WePM as a promising foundation for capable on-device AI under tight memory constraints.

Our contributions are as follows:
\begin{itemize}[
  leftmargin=1.5em,
  labelsep=0.55em,
  itemsep=4pt,
  topsep=4pt,
  parsep=0pt
]
\item We introduce and formalize heterogeneous capability density, a new systems-level objective for on-device AI that jointly captures capability breadth, balanced task quality, and resident inference memory. HCD reframes efficient on-device intelligence from optimizing isolated tasks or model size to maximizing the useful heterogeneous capabilities delivered per unit of memory.
\item We formulate the unification of representation, generation, and decision in terms of heterogeneous evidence requirements and shared-parameter optimization interference, exposing coupled conflicts in visual granularity, cross-task supervision, and effective update magnitude. We address them through MGVB, ULKD, SEMO, and task-appropriate output paths.
\item We evaluate WePM across 30 embedding, 39 generation, and seven typed-decision benchmarks. In the joint representation--generation setting shared with existing unified methods, it leads all five evaluation groups and raises the Joint score from the strongest baseline's 38.15 to 52.74. WePM-4B also ranks first in average single-task performance across all three capability families. Compared with the separate-model stack with the highest HCD, WePM achieves slightly higher balanced performance with 61.9\% less BF16 resident weight memory, yielding 2.64 times the HCD.
\end{itemize}

\section{Unified Multimodal Models: Formulation and Tensions}
\label{sec:problem}

\subsection{Existing Formulations}
\label{sec:existing-formulations}

A natural way to combine different tasks such as  retrieval and generation is to reuse a generative backbone, define an additional embedding readout, and train the two task losses with a single weighted-sum objective. We take this shared-backbone construction as the baseline formulation of this paper: it is the common recipe behind prior unified models, the substrate that WePM retains, and the reference configuration in our experiments. Existing methods instantiate the representation readout differently: GRIT uses mean-pooled input states~\cite{muennighoff2025grit}, whereas VLM2Vec and E5-V extract last-token states for multimodal embedding~\cite{jiang2025vlm2vec,jiang2024e5v}. Our baseline follows the latter design, projecting the final hidden state after multimodal prefill without inserting an additional representation token, while retaining standard causal decoding for generation.

More specifically, we consider retrieval examples $(q,c^+,\{c_j^-\})\sim\mathcal D_{\mathrm{emb}}$ and generation examples $(x,y)\sim\mathcal D_{\mathrm{gen}}$, where an input may contain an image, text, or both. The baseline serves both tasks with a single shared backbone $F_\theta$, rather than maintaining separate retrieval and generation models. Let $V(I)$ and $T(\cdot)$ denote the visual and text tokens of an input, respectively. For an embedding input $u$, the backbone prefills the sequence $z_u=[V(I_u);T(u)]$. If $m(u)$ denotes its final valid input position, retrieval projects the corresponding final-layer hidden state into a normalized embedding,
\[
e_\theta(u)=\operatorname{norm}\!\left(W_e [F_\theta(z_u)]_{m(u)}\right),\qquad W_e\in\mathbb R^{d_e\times d}.
\]
With a candidate set $\mathcal C(q)$ that contains $c^+$, its sampled hard negatives $\{c_j^-\}$, and the other in-batch candidates of the same global batch, with candidates duplicating the positive masked out, define the temperature-scaled similarity $s_\theta(q,c)=\operatorname{sim}(e_\theta(q),e_\theta(c))/\tau$. The per-query list-wise contrastive loss then takes the equivalent log-sum-exp form
\[
\mathcal L_{\mathrm{emb}}(q)
=\log\!\sum_{c\in\mathcal C(q)}\exp\bigl(s_\theta(q,c)\bigr)
 -s_\theta(q,c^+).
\]
Generation uses the backbone's standard causal language-modeling path, without an additional representation token, and predicts the answer autoregressively with the per-token loss
\[
\mathcal L_{\mathrm{gen}}(x,y)=-|y|^{-1}\sum_{t=1}^{|y|}\log p_\theta(y_t\mid I,x,y_{<t}).
\]
The two paths therefore share the vision encoder and all backbone parameters but use task-appropriate outputs: retrieval stops after multimodal prefill and applies $W_e$ once, whereas generation applies the language-modeling head throughout autoregressive decoding. For training, the baseline combines the two task losses through a weighted objective, $\mathcal L=\lambda_{\mathrm{emb}}\mathcal L_{\mathrm{emb}}+\lambda_{\mathrm{gen}}\mathcal L_{\mathrm{gen}}$. This shared-backbone formulation enables compact joint training, but simply combining the task losses leaves visual-evidence allocation and interference between shared-parameter updates unresolved, as discussed in Section~\ref{sec:shared-capacity-tensions}.

\subsection{Conflicts Under Shared Capacity}
\label{sec:shared-capacity-tensions}

Unifying heterogeneous tasks---such as representation, generation, and decision---within a compact shared backbone introduces two fundamental difficulties. First, the three capabilities require visual evidence at different spatial granularities, despite sharing the same limited visual-token budget. Second, their training data are typically collected in separate task-specific corpora. Existing unified training recipes therefore commonly alternate among task-specific batches, causing consecutive updates to optimize different output semantics over the same backbone parameters. We trace these two difficulties to a granularity conflict in the retained evidence and an optimization conflict in the shared parameter updates.

\themebf{Granularity conflict}---which visual evidence should be retained? Representation may rely on global semantics for broad retrieval, but distinguishing hard negatives often requires regional attributes and fine details. Generation similarly combines scene-level context with localized evidence for OCR and detail-centric question answering, while decision may depend on either global state or local text, values, and regions specified by its schema. Because all three capabilities operate under a visual-token budget of at most $B$, uniformly compressing an image can preserve evidence useful to one task while discarding evidence required by another. The first challenge is therefore to preserve task-relevant evidence at multiple granularities within a single fixed token budget.

The second difficulty arises during optimization. Let $\mathcal T=\{\mathrm{emb},\mathrm{gen},\mathrm{dec}\}$ and $g_i=\nabla_\theta\mathcal L_i$ denote the gradient of task $i$ with respect to the shared parameters. Under task-alternating training, an update using task $i$ changes the loss of another task $j$ according to
\[
\mathcal L_j(\theta-\alpha g_i)
=
\mathcal L_j(\theta)
-\alpha\langle g_j,g_i\rangle
+O(\alpha^2),
\]
for sufficiently small $\alpha>0$. When $\langle g_j,g_i\rangle<0$, improving task $i$ locally increases the loss of task $j$, yielding direct directional interference. Even when the inner product is nonnegative, unequal gradient norms and update frequencies can allow one task to dominate the shared parameters. There are two natural approaches to mitigating this interference, but each exposes a distinct challenge.

\themebf{Supervision conflict}---can task-specific data be converted into paired supervision? One possible approach is to annotate every input with targets for all capabilities, so that representation, generation, and decision can be optimized from semantically aligned examples. Constructing such data at scale, however, is expensive and often fundamentally ambiguous. For example, a representation query such as ``a small red flower'' specifies a meaningful retrieval target but does not determine a unique generation response. Both ``a small red flower blooms on the hillside'' and ``a small red flower grows in the woods'' are plausible continuations. Assigning either response as the sole target introduces an arbitrary interpretation rather than genuine cross-task supervision. More generally, relevance relations, open-ended sequences, and typed decision labels define different target spaces and cannot always be converted into one another through hard annotations. The supervision challenge is therefore to obtain complementary cross-task learning signals without requiring exhaustive paired data or imposing a unique target where none exists.

\themebf{Magnitude conflict}---which task determines the shared update? An alternative to constructing fully paired data is to integrate gradients from different task-specific batches into one logical update. However, gradient integration alone does not ensure balanced task influence. For $n_i$ batches of task $i$, coefficient $\lambda_i$, and mean gradient $\bar g_i$, the resulting update is
\[
g_{\mathrm{joint}}
=
\sum_{i\in\mathcal T}
\lambda_i\sum_{b=1}^{n_i}\nabla_\theta\ell_{i,b}
=
\sum_{i\in\mathcal T}n_i\lambda_i\bar g_i.
\]
The effective contribution of task $i$ is therefore determined jointly by its native gradient magnitude, its coefficient $\lambda_i$, and its exposure frequency $n_i$. These quantities are not naturally comparable across heterogeneous objectives: representation losses are typically reduced over queries or candidate sets, generation losses over supervised tokens, and decision losses over typed instances. Their data pools may also differ substantially in size, causing some tasks to be sampled more frequently than others. Consequently, equal coefficients do not imply equal influence, and equal exposure does not imply balanced updates. A task with a larger effective contribution can dominate the joint gradient and overwrite capabilities learned from other tasks, even when their gradient directions are not directly opposed. Conversely, suppressing that contribution too aggressively can prevent the task from learning sufficiently. The magnitude conflict therefore lies in the mismatch between each task's numerically induced optimization strength and its intended importance in the final capability trade-off. The challenge is to control both task weighting and task exposure so that the joint update reflects a deliberate balance rather than incidental differences in loss normalization, annotation density, or corpus size.

These conflicts motivate the three corresponding components of WePM. MGVB reconciles heterogeneous evidence requirements within a shared visual-token budget. ULKD supplies cross-capability soft supervision without requiring unique hard targets or exhaustively paired annotations. SEMO controls the effective contribution of each task through explicit loss coefficients and task exposure, while using memory-efficient sequential accumulation to realize the specified joint update.

\section{WePM: A Pocket Model for Heterogeneous On-Device Intelligence}
\label{sec:method}

\begin{figure*}[!h]
  \centering
  \includegraphics[width=\textwidth]{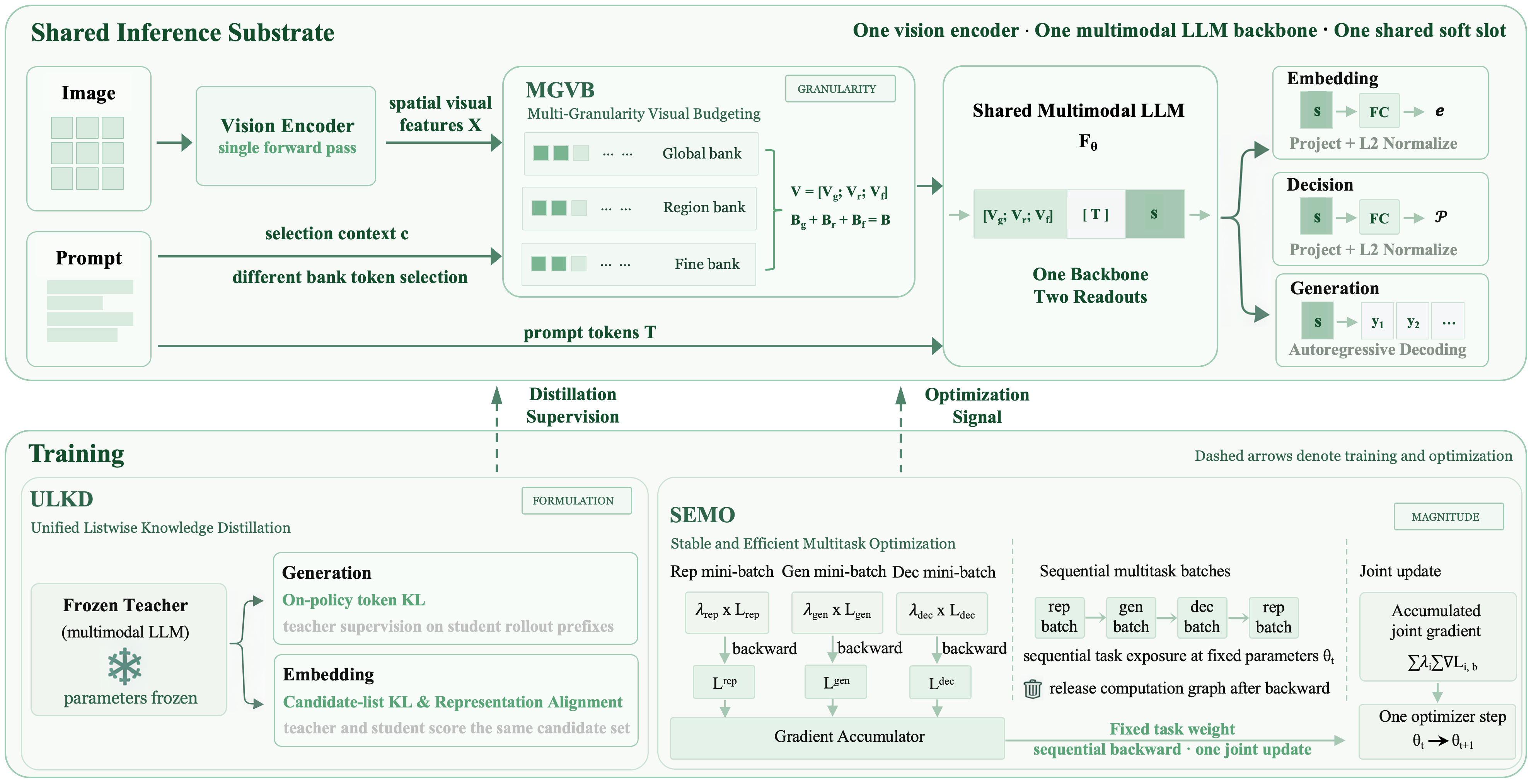}
  \caption{Overview of WePM. A shared vision encoder and multimodal backbone produce task-appropriate outputs: normalized embeddings for representation, autoregressive responses for generation, and schema-conditioned scores for typed decision. MGVB allocates a fixed visual-token budget across global, regional, and fine-grained banks using the available textual context. During training, ULKD supplements representation and generation data with token-level generation guidance, candidate-list preferences, and representation alignment from capability-specific teachers. SEMO controls each task’s contribution through explicit loss weights and exposure schedules, while sequential gradient accumulation executes the joint update within training-memory constraints.}
  \label{fig:wepm-architecture}
\end{figure*}

\subsection{Multi-Granularity Visual Budgeting} 

Representation, generation, and visually grounded decision require evidence at different spatial scales. Coarse-grained retrieval is largely governed by global scene semantics, whereas discriminating hard negatives often requires object-level and relational cues. OCR and detail-centric visual question answering depend on localized evidence that may occupy only a few image patches, while a decision may depend on either global state or a local value specified by its criteria. Under a fixed visual-token budget, uniform compression inevitably preserves redundant information at some scales while discarding evidence critical at others. We therefore introduce the \emph{Multi-Granularity Visual Budgeting}, which explicitly distributes the available visual capacity across global, regional, and fine-grained representations.

Given spatially indexed features $X=\{x_j\}_{j=1}^{N}$ produced by a single vision-encoder forward pass, MGVB constructs three complementary token banks:
\[
V=[V_g;V_r;V_f],\qquad B_g+B_r+B_f=B,
\]
where $B_g$, $B_r$, and $B_f$ denote the token budgets of the global, regional, and fine-grained banks, respectively, and $B$ is the total number of visual tokens supplied to the language backbone. The global bank employs learned latent queries to aggregate image-level semantics. The regional bank summarizes candidate regions, preserving object identities, spatial layouts, and inter-object relations. The fine-grained bank retains selected patch features that capture small objects, scene text, and other localized details that would otherwise be suppressed by regional aggregation. Together, these banks provide a compact yet scale-complementary visual representation.

While the three banks determine \emph{which scales} of evidence are retained, the accompanying text determines \emph{which content} is most relevant. Accordingly, MGVB conditions token selection in the regional and fine-grained banks on the available textual context $c$. Let $R_k$ be a candidate region with pooled representation $r_k$, and let $x_j$ be a patch feature associated with region $R_{k(j)}$. Their relevance scores are computed as
\[
a_k^{r}=f_r(r_k,x_{\mathrm{global}},c),
\qquad
a_j^{f}=f_f(x_j,r_{k(j)},c),
\]
where $x_{\mathrm{global}}$ denotes the global image representation, and $f_r$ and $f_f$ are learnable scoring functions. The top-ranked regions and patches are retained under budgets $B_r$ and $B_f$, respectively. For instance, a question concerning a small chart label favors the corresponding local patches, whereas a question about the overall trend emphasizes the chart structure. Importantly, this context-aware selection only reallocates capacity within a shared representation; it introduces neither task-specific visual encoders nor separate backbone branches.

MGVB follows the same selection principle across the three capabilities whenever visual input is present. The context $c$ is instantiated by the text accompanying the image, such as a generation instruction, a retrieval query or caption, or a decision schema and its criteria. If no text is available, MGVB uses a query-independent context $c_{\emptyset}$, ensuring that indexed representations do not depend on future queries. Finally, each retained token is augmented with granularity and spatial-position embeddings to encode its scale and spatial origin before being passed to the shared backbone. 

\subsection{Unified Listwise Knowledge Distillation}
\label{sec:ulkd}

Retrieval and generation corpora provide complementary but typically unpaired forms of supervision. Retrieval examples contain relevance annotations but no reference responses, whereas generation examples provide target sequences without candidate-level relevance labels. A straightforward way to bridge this gap would be to annotate each example for both capabilities. As discussed in Section~\ref{sec:shared-capacity-tensions}, however, such cross-capability annotation is not only expensive but can also be fundamentally ambiguous. For example, a retrieval query such as ``a small red flower'' specifies meaningful relevance relations but admits many valid continuations rather than a unique generation target. Assigning one pseudo-response as the ground truth would impose an arbitrary interpretation on the query.

To address this supervision conflict, we introduce \emph{Unified Listwise Knowledge Distillation}, which transfers soft teacher preferences instead of constructing unique hard targets for missing annotations. ULKD retains the native supervision available in each corpus while supplementing partially annotated inputs with capability-specific distributional guidance. In the present formulation, ULKD bridges the representation and generation supervision spaces; the decision path continues to use its native schema-conditioned typed objective.

ULKD places retrieval and generation distillation within a common distribution-matching framework using two frozen capability-specific teachers. At each generation step, the choice set is the vocabulary, $\mathcal A_{\mathrm{gen}}=\mathcal V$, whereas retrieval defines a distribution over the candidate list, $\mathcal A_{\mathrm{emb}}=\mathcal C(q)$. In both cases, the teacher provides relative preferences over the corresponding choice set, and the student minimizes a KL divergence to the teacher distribution. This formulation preserves multiple plausible alternatives instead of collapsing supervision into a single hard token, response, or positive candidate.

For generation, directly assigning a pseudo-response to an open-ended retrieval query would impose one arbitrary interpretation as the target. ULKD instead performs on-policy distillation along trajectories sampled from the student itself~\cite{thinkingmachines2025opd}. Let $\mathcal D_{\mathrm{emb}}^{q}$ denote the query marginal of the retrieval corpus. For each query prompt $q$, the student samples a continuation $\hat y\sim p_S(\cdot\mid q)$, and the generation teacher evaluates every visited prefix. Following the cited formulation, we minimize the per-token reverse KL, with the student distribution as its first argument:
\[
\mathcal L_{\mathrm{opd}}=
\mathbb E_{q\sim\mathcal D_{\mathrm{emb}}^{q},\,\hat y\sim p_S(\cdot\mid q)}\!\left[
\frac{1}{|\hat y|}\sum_{t=1}^{|\hat y|}
D_{\mathrm{KL}}\!\left(
 p_S(\cdot\mid q,\hat y_{<t})\,\Vert\,
 p_T(\cdot\mid q,\hat y_{<t})
\right)\right].
\]
Unlike sequence-level pseudo-labeling, this objective does not designate any complete continuation as uniquely correct. Instead, it aligns the student's token distribution with the teacher on states induced by the student's own policy, thereby providing generation supervision precisely along the trajectories encountered at inference time.

For retrieval, the embedding teacher and student score the same candidate set $\mathcal C(q)$. Their listwise distributions can be written together as
\[
r_A(c\mid q)=\operatorname{softmax}_{c\in\mathcal C(q)}\!\bigl(s_A(q,c)/\tau_A\bigr),
\qquad A\in\{T,S\},
\]
where $\tau_T$ and $\tau_S$ are temperature parameters. ULKD transfers both the teacher's relative ordering over candidates and its instance-level representation. Specifically, a learnable projection $P$ maps teacher embeddings into the student space:
\[
\begin{aligned}
\mathcal L_{\mathrm{rank}}&=\mathbb E_{q,\mathcal C(q)}
D_{\mathrm{KL}}\!\left(r_T(\cdot\mid q)\,\Vert\,r_S(\cdot\mid q)\right),\\
\mathcal L_{\mathrm{repr}}&=\mathbb E_u\bigl\|e_S(u)-P(e_T(u))\bigr\|_2^2,\\
\mathcal L_{\mathrm{ULKD}}&=\eta_{\mathrm{opd}}\mathcal L_{\mathrm{opd}}+
\eta_{\mathrm{rank}}\mathcal L_{\mathrm{rank}}+
\eta_{\mathrm{repr}}\mathcal L_{\mathrm{repr}},
\end{aligned}
\]
where $\eta_{\mathrm{opd}}$, $\eta_{\mathrm{rank}}$, and $\eta_{\mathrm{repr}}$ control the three distillation terms.

During training, ULKD forms an auxiliary supervision stream alongside the original generation and retrieval objectives, while decision remains optimized with its native typed loss. Generation examples retain their standard token-level supervision, and retrieval examples retain their hard contrastive supervision through the last-state embedding readout. For a retrieval query, ULKD additionally samples an on-policy continuation and obtains token-level guidance from the generation teacher, while the embedding teacher transfers candidate-level ranking preferences and instance-level representations. ULKD therefore enriches task-specific data with complementary soft supervision without replacing their native objectives, requiring exhaustively paired annotations, or imposing a unique response on an open-ended query.

\subsection{Stable and Efficient Multitask Optimization}

The magnitude conflict identified in Section~\ref{sec:shared-capacity-tensions} arises because the effective influence of each task depends jointly on its native gradient scale, loss coefficient, and exposure frequency. A multitask optimizer must therefore determine not only how strongly each task is weighted, but also how often it contributes to the shared update. Adaptive weighting methods such as Fast Adaptive Multitask Optimization (FAMO)~\cite{liu2023famo} adjust task coefficients according to recent loss reduction. For heterogeneous objectives, however, short-term loss progress need not reflect the desired trade-off in downstream capability, and estimating such progress requires post-update loss re-evaluation and additional forward passes. Moreover, adaptive coefficients regulate only one component of task influence; the exposure assigned to each task remains a separate design choice. We therefore introduce \emph{Stable and Efficient Multitask Optimization}, which explicitly specifies both task weighting and task exposure while executing the resulting joint update through memory-efficient sequential gradient accumulation.

Formally, let $\mathcal T=\{\mathrm{emb},\mathrm{gen},\mathrm{dec}\}$ denote the representation, generation, and decision tasks. For task $i\in\mathcal T$, let $\mathcal B_{i,t}$ be the set of batches assigned to logical update $t$, with $n_{i,t}=|\mathcal B_{i,t}|$ determined by a predefined exposure schedule derived from the task-pool sizes. For each batch $b$, $\ell_{i,b}$ denotes the task-native loss after the corresponding per-query, per-token, or per-decision-instance reduction. SEMO defines the logical-update objective as
\[
\mathcal L_t^{\mathrm{SEMO}}(\theta)
=
\sum_{i\in\mathcal T}
\lambda_i
\sum_{b\in\mathcal B_{i,t}}
\ell_{i,b}(\theta),
\qquad \lambda_i>0,
\]
where $\lambda_i$ remains fixed throughout training. Defining
\[
\bar g_{i,t}
=
\frac{1}{n_{i,t}}
\sum_{b\in\mathcal B_{i,t}}
\nabla_\theta\ell_{i,b}(\theta_t),
\]
the contribution of task $i$ to the joint update is $n_{i,t}\lambda_i\bar g_{i,t}$. SEMO therefore makes the two controllable factors governing task influence explicit: $\lambda_i$ determines the per-batch weight, while $n_{i,t}$ determines the task exposure. Both are specified deliberately rather than being implicitly determined by loss normalization or incidental sampling frequency.

A simultaneous implementation of this objective would materialize the losses for all representation, generation, and decision batches before backpropagation, retaining their computation graphs at the same time. This execution strategy exceeds the available training memory in our setting. SEMO instead processes the constituent batches sequentially. Each batch performs a forward pass followed immediately by weighted backpropagation, after which its computation graph is released. The shared parameters remain fixed at $\theta_t$ until all batches assigned to logical update $t$ have contributed their gradients. Consequently,
\[
g_t^{\mathrm{seq}}
=
\sum_{i\in\mathcal T}
\lambda_i
\sum_{b\in\mathcal B_{i,t}}
\nabla_\theta\ell_{i,b}(\theta_t)
=
\nabla_\theta
\mathcal L_t^{\mathrm{SEMO}}(\theta_t).
\]
Sequential accumulation therefore computes the gradient of the specified joint objective while requiring memory for only one batch graph at a time.

SEMO separates the statistical design of the joint update---task coefficients and exposure frequencies---from its memory-efficient execution. It does not project or otherwise modify task-gradient directions; instead, it addresses the magnitude conflict by ensuring that each task's effective contribution follows an explicit capability trade-off rather than uncontrolled differences in loss scale, annotation density, or corpus size. Section~\ref{sec:ablation-studies} compares this fixed and exposure-aware design with adaptive FAMO variants and evaluates the memory efficiency of sequential execution.

\section{Experiments}
\label{sec:experiments}

\subsection{Experimental Details}
\label{sec:experimental-details}

We initialize WePM from the pretrained Qwen3.5-4B/2B dense model~\cite{qwen3.5} and fine-tune all parameters. Unless otherwise specified, \emph{WePM} refers to this 4B model throughout the paper. We choose this scale because deployment profiling identifies it as approximately the practical upper bound for dense models on representative on-device platforms such as Apple M3: larger models degrade interactive performance and user experience. The maximum sequence length during training is 8,192 tokens, and the combined generation, embedding, and decision corpus contains approximately 100 million cleaned samples.

Retrieval instances combine negatives from their original datasets with additional hard negatives mined using the WeMM-9B embedding model~\cite{zhou2026wemm}. After negative expansion and cross-device aggregation, embedding training uses an effective global batch size of approximately 20k, exposing the model to more confusable candidates during contrastive learning. For ULKD, Qwen3.5-27B Dense~\cite{qwen3.5} provides generation-side supervision, while WeMM-9B embedding~\cite{zhou2026wemm} provides embedding-side supervision. These capability-specific teachers guide the shared student through their corresponding generation and representation objectives.

During full-capability training, generation, embedding, and decision batches are sampled in a 7:5:1 ratio according to the data distribution. Their training losses are reduced per supervised token, per query, and per typed decision instance, respectively. The controlled representation--generation weighting ablations use a separate schedule of three embedding and five generation batches per logical update. The weights in Table~\ref{tab:semo-weighting} denote per-batch loss coefficients ($\lambda_{\mathrm{emb}}/\lambda_{\mathrm{gen}}$), not task-exposure ratios; $B$ in Table~\ref{tab:mgvb-budget} denotes the MGVB visual-token budget.

For comparisons with existing unified methods and controlled ablations, we use a shared-capability protocol covering 16 representation and generation benchmarks in five groups. Image retrieval (ImageRet.) comprises MSCOCO image-to-text retrieval~\cite{lin2014coco}, CIRR~\cite{liu2021cirr}, and VisDial~\cite{das2017visualdialog}. Text retrieval (TextRet.) comprises ArguAna~\cite{wachsmuth2018arguana}, FiQA-2018~\cite{maia2018fiqa}, CQADupStack-Gaming~\cite{hoogeveen2015cqadupstack}, and SCIDOCS~\cite{cohan2020specter} from BEIR~\cite{thakur2021beir}. Mathematical and scientific reasoning (Math) comprises MMMU~\cite{yue2023mmmu}, MathVista~\cite{lu2023mathvista}, and MathVision~\cite{wang2024mathvision}. Visual question answering (VQA) comprises MM-Vet~\cite{yu2023mmvet}, AI2D~\cite{kembhavi2016ai2d}, and VSR-ZeroShot~\cite{liu2022vsr}. OCR and document understanding (OCR) comprises OCRBench~\cite{liu2023ocrbench}, ChartQA~\cite{masry2022chartqa}, and TextVQA~\cite{singh2019textvqa}. We reproduce all baseline results using the same input resolution and evaluation scripts as WePM.

For both evaluation protocols, each benchmark score is reported using the benchmark's standard evaluation metric. Each group score is the arithmetic mean of its benchmark scores. The Embedding score averages ImageRet. and TextRet., the Generation score averages Math, VQA, and OCR, and Joint averages the Embedding and Generation scores, giving the two capabilities equal weight. Higher scores are better throughout. The full-capability protocol separately evaluates representation, generation, decision, resident memory, and HCD. The embedding evaluation includes text tasks from MTEB~\cite{muennighoff2022mteb}, with their individual dataset sources cited in Table~\ref{tab:single-task-embedding}.

\subsection{Comparison with Models on Single Task}
\label{sec:single_sota}

We compare WePM with specialized models on generation, embedding, and typed-decision benchmarks. Although WePM uses one shared 4B backbone rather than a capability-specific model, it achieves strong average performance across all three evaluations, suggesting that joint capability learning can preserve task-level competitiveness.

\themebf{Generation.} As shown in Table~\ref{tab:single-task-generation}, WePM-4B obtains the best overall average of 62.41, outperforming the strongest baseline, Qwen3.5-4B, by 0.58 points. It ranks first on 14 of the 39 benchmarks and second on another 8, with leading results spanning visual perception, factual VQA, embodied reasoning, and knowledge-intensive reasoning. The gains are especially pronounced on VLM's Are Blind (+12.45 over the runner-up), GPQA Diamond (+5.56), and SimpleVQA (+4.14), indicating that the unified training improves both fine-grained visual grounding and difficult reasoning rather than specializing in a narrow benchmark family. The 2B variant, WePM-2B, reaches 55.69 with half the parameters, exceeding the two 2B baselines by 11.20 and 16.42 points and surpassing three of the four 4B-scale baselines.

\themebf{Embedding.} Table~\ref{tab:single-task-embedding} shows that WePM reaches an average of 72.86, narrowly exceeding the specialized Qwen3-VL-Embedding-8B model by 0.04 points. WePM ranks first on four benchmarks and second on two, and is particularly strong on semantic retrieval and clustering tasks. On AskUbuntuDupQuestions, it achieves 64.29, exceeding the strongest baseline by 20.85 points; it also leads on SCIDOCS, Banking77, and TwentyNewsgroups. The 2B variant, WePM-2B, reaches 70.73, matching the specialized Qwen3-VL-Embedding-2B (70.73) at the same scale and trailing WePM-4B by 2.13 points. These results show that the shared generative backbone can retain highly competitive representation quality without a dedicated embedding-only architecture.

\themebf{Decision.} As reported in Table~\ref{tab:single-task-decision}, WePM achieves the highest average score of 61.11, slightly surpassing d3-mini-4.5B at 61.06. It obtains the best results on JevAny (MM) and Valen-General5k, ranks second on Open-Jev, and improves Valen-General5k by 8.88 points over the strongest external model. The 2B variant, WePM-2B, already reaches 55.81 and ranks fourth overall, above several 4B- and 9B-scale decision models. Together with the generation and embedding results, this confirms that WePM can incorporate a native typed-decision capability while preserving strong general-purpose generation and representation performance.

\begin{table*}[t]
\centering
\caption{Comparison on single-task generation benchmarks. Models are arranged as columns and sorted by Average in ascending order, with the two WePM variants listed last. The best result for each benchmark is highlighted and the second-best result is underlined; higher is better.}
\label{tab:single-task-generation}
\setlength{\tabcolsep}{1.5pt}
\renewcommand{\arraystretch}{0.92}
\scriptsize
\begin{tabularx}{\textwidth}{@{}>{\raggedright\arraybackslash}m{0.24\textwidth}*{8}{>{\centering\arraybackslash}m{\dimexpr 0.0755\textwidth-3pt\relax}}>{\cellcolor{jcolour!30}\centering\arraybackslash}m{\dimexpr 0.0755\textwidth-3pt\relax}>{\cellcolor{jcolour!30}\centering\arraybackslash}m{\dimexpr 0.0755\textwidth-3pt\relax}@{}}
\specialrule{\heavyrulewidth}{0pt}{0pt}
\rowcolor{jcolour!15}Benchmark & \shortstack{SmolVLM2\\2B~\cite{huggingface2025smolvlm2}} & \shortstack{Gemma-4\\E2B-it~\cite{gemmateam2026gemma4}} & \shortstack{Phi-3.5V\\4B~\cite{microsoft2024phi35vision}} & \shortstack{Gemma-4\\E4B-it~\cite{gemmateam2026gemma4}} & \shortstack{MiniCPM-V\\4.0~\cite{openbmb2025minicpmv4}} & \shortstack{InternVL3\\8B~\cite{zhu2025internvl3}} & \shortstack{MiniCPM-o\\4.5~\cite{openbmb2026minicpmo45}} & \shortstack{Qwen3.5\\4B~\cite{qwen3.5}} & \cellcolor{jcolour!30}\themebf{\shortstack{WePM\\2B}} & \cellcolor{jcolour!30}\themebf{\shortstack{WePM\\4B}} \\
\specialrule{\lightrulewidth}{0pt}{0pt}
\rowcolor{gray!8} AI2D~\cite{kembhavi2016ai2d} & 64.18 & 71.47 & 78.95 & 78.89 & 80.70 & 85.40 & 86.82 & \themebf{87.56} & 82.45 & \underline{87.24} \\
MMStar~\cite{chen2024mmstar} & 39.33 & 46.60 & 49.93 & 54.53 & 61.93 & 70.47 & \themebf{71.53} & \underline{71.13} & 61.47 & 68.13 \\
\rowcolor{gray!8} MMVet~\cite{yu2023mmvet} & 22.48 & 24.31 & \themebf{62.84} & 28.44 & 30.73 & \underline{43.12} & 34.40 & 38.07 & 31.65 & 35.78 \\
POPE~\cite{li2023pope} & 89.10 & 87.13 & 90.19 & 89.12 & 91.50 & \underline{92.30} & 91.48 & 91.81 & \underline{92.30} & \themebf{92.88} \\
\rowcolor{gray!8} VSR-ZeroShot~\cite{liu2023vsr} & 70.21 & 61.21 & 59.41 & 69.48 & 77.91 & 83.88 & 82.16 & \underline{84.04} & 75.45 & \themebf{85.92} \\
HallusionBench~\cite{guan2024hallusionbench} & 58.02 & 61.03 & 64.39 & 65.72 & 64.30 & 64.57 & \themebf{72.90} & 67.85 & 67.05 & \underline{72.28} \\
\rowcolor{gray!8} SimpleVQA~\cite{ji2025simplevqa} & 10.07 & 13.33 & 11.96 & 13.83 & 15.24 & 17.53 & 20.59 & \underline{23.51} & 18.81 & \themebf{27.65} \\
RealWorldQA~\cite{xai2024realworldqa} & 57.12 & 48.10 & 58.30 & 56.08 & 63.40 & 67.71 & 67.32 & \underline{73.59} & 65.23 & \themebf{73.86} \\
\rowcolor{gray!8} VLM's Are Blind~\cite{rahmanzadehgervi2024vlmsblind} & 32.52 & 44.84 & 40.96 & 60.09 & 45.02 & 50.65 & 64.92 & 66.04 & \underline{74.94} & \themebf{78.49} \\
LingoQA~\cite{marcu2024lingoqa} & 12.00 & 7.40 & \themebf{53.40} & 12.30 & 11.50 & 13.40 & 15.60 & 13.10 & \underline{18.60} & 17.90 \\
\rowcolor{gray!8} BabyVision~\cite{chen2026babyvision} & 11.60 & 11.60 & 12.11 & 10.57 & \underline{17.27} & 12.89 & 13.66 & 13.14 & 16.24 & \themebf{17.53} \\
ChartQA~\cite{masry2022chartqa} & 52.24 & 57.92 & 74.44 & 63.72 & 82.04 & \underline{86.20} & 84.56 & \themebf{89.80} & 83.60 & 86.08 \\
\rowcolor{gray!8} OCRBench~\cite{liu2023ocrbench} & 63.20 & 71.50 & 64.60 & 76.20 & \underline{87.00} & \themebf{91.10} & 85.60 & 86.10 & 82.90 & 85.70 \\
TextVQA~\cite{singh2019textvqa} & 68.26 & 66.90 & 55.44 & 69.72 & 56.45 & \underline{86.30} & 82.88 & 57.68 & 85.14 & \themebf{88.08} \\
\rowcolor{gray!8} CharXiv~\cite{wang2024charxiv} & 27.52 & 34.20 & \underline{46.74} & 38.20 & 45.26 & 44.60 & 45.04 & \themebf{52.20} & 41.02 & 42.56 \\
MMLongBench-Doc~\cite{ma2024mmlongbench} & 15.69 & 19.61 & \underline{33.79} & 22.62 & 23.23 & 33.48 & \underline{33.79} & \themebf{36.65} & 23.38 & 32.13 \\
\rowcolor{gray!8} OCRBench-V2~\cite{fu2025ocrbenchv2} & 31.66 & 43.01 & 34.62 & 48.59 & 56.35 & \underline{60.07} & 59.68 & \themebf{62.51} & 50.04 & 59.50 \\
MathVista~\cite{lu2023mathvista} & 46.00 & 41.00 & 48.60 & 49.50 & 64.10 & 71.00 & \underline{80.20} & \themebf{81.00} & 60.40 & 68.90 \\
\rowcolor{gray!8} MMMU~\cite{yue2023mmmu} & 34.00 & 39.05 & 40.67 & 49.62 & 48.00 & 58.86 & \themebf{63.33} & \underline{60.86} & 46.67 & 54.67 \\
TallyQA~\cite{acharya2019tallyqa} & 69.50 & 63.46 & 71.70 & 65.66 & 77.09 & \underline{81.14} & 76.97 & 79.08 & 75.72 & \themebf{81.36} \\
\rowcolor{gray!8} CountBench~\cite{paiss2023countbench} & 46.20 & 23.41 & 66.53 & 52.36 & 82.96 & 81.93 & \themebf{87.27} & 84.19 & 82.96 & \underline{84.80} \\
ERQA~\cite{team2025geminirobotics} & 31.50 & 31.75 & 30.25 & 34.50 & 38.25 & 37.25 & 40.50 & \underline{43.75} & 39.25 & \themebf{45.25} \\
\rowcolor{gray!8} EmbSpatialBench~\cite{du2024embspatial} & 38.24 & 46.65 & 55.63 & 64.48 & 60.80 & 75.49 & 70.14 & \underline{75.71} & 71.35 & \themebf{76.87} \\
MMBench-EN-DEV~\cite{liu2023mmbench} & 71.94 & 76.07 & 79.90 & 82.26 & 85.11 & 89.60 & \underline{91.04} & \themebf{91.41} & 87.24 & 90.61 \\
\rowcolor{gray!8} MMBench-EN-TEST~\cite{liu2023mmbench} & 73.24 & 76.48 & 79.87 & 83.48 & 84.98 & 86.51 & \underline{90.05} & \themebf{90.57} & 85.22 & 88.92 \\
MMBench-CN-TEST~\cite{liu2023mmbench} & 60.23 & 75.57 & 73.34 & 81.39 & 83.08 & 86.61 & \underline{88.42} & \themebf{89.25} & 83.08 & 87.07 \\
\rowcolor{gray!8} SLAKE~\cite{liu2021slake} & 29.44 & 38.11 & 38.44 & 39.49 & 27.84 & 44.50 & 45.12 & \themebf{57.50} & 41.07 & \underline{53.74} \\
PMC-VQA~\cite{zhang2023pmcvqa} & 39.88 & 51.48 & 41.70 & 44.22 & 53.50 & 53.08 & 54.19 & \themebf{60.50} & 54.07 & \underline{58.79} \\
\rowcolor{gray!8} MedXpertQA-MM~\cite{zuo2025medxpertqa} & 21.10 & 23.35 & 21.70 & 27.75 & 21.00 & 23.00 & 25.60 & \themebf{36.05} & 23.90 & \underline{27.85} \\
MedXpertQA-Text~\cite{zuo2025medxpertqa} & 9.92 & 14.61 & 10.65 & \themebf{18.45} & 7.71 & 13.76 & 13.71 & \underline{17.76} & 14.16 & 16.24 \\
\rowcolor{gray!8} C-Eval~\cite{huang2023ceval} & 28.59 & 48.48 & 38.49 & 53.01 & 57.33 & \themebf{81.58} & 73.61 & 71.85 & 62.87 & \underline{75.23} \\
MMLU-Redux~\cite{gema2024mmluredux} & 36.27 & 57.87 & 55.83 & 56.57 & 57.30 & 74.53 & \underline{74.93} & 71.40 & 65.43 & \themebf{77.20} \\
\rowcolor{gray!8} MMLU-Pro~\cite{wang2024mmlupro} & 17.17 & 44.61 & 33.47 & 55.11 & 37.67 & 58.12 & \underline{63.66} & 59.69 & 47.86 & \themebf{66.00} \\
MMMLU~\cite{openai2024mmmlu} & 40.26 & 66.94 & 62.51 & 65.70 & 60.37 & 76.29 & \themebf{78.22} & 74.17 & 65.87 & \underline{78.01} \\
\rowcolor{gray!8} Global-PIQA~\cite{singh2025globalpiqa} & 25.53 & 44.41 & 28.67 & \themebf{54.40} & 28.34 & 43.47 & \underline{48.42} & 45.03 & 36.49 & 46.21 \\
SuperGPQA~\cite{duan2025supergpqa} & 12.14 & 20.82 & 18.49 & 14.77 & 18.36 & 29.68 & 33.54 & \underline{36.13} & 23.59 & \themebf{38.27} \\
\rowcolor{gray!8} GPQA Diamond~\cite{rein2023gpqa} & 27.27 & 31.31 & 34.85 & \underline{48.48} & 23.74 & 36.87 & 38.89 & 44.44 & 35.86 & \themebf{54.04} \\
IFEval~\cite{zhou2023ifeval} & 28.90 & 49.16 & 32.73 & 50.96 & 47.12 & 46.04 & \themebf{56.12} & \underline{55.76} & 39.45 & 51.44 \\
\rowcolor{gray!8} BFCL v3~\cite{yan2024bfcl} & 19.20 & 0.40 & 62.30 & 0.60 & 46.30 & \underline{67.60} & 63.90 & \themebf{70.60} & 59.10 & 60.80 \\
\specialrule{\lightrulewidth}{0pt}{0pt}
Ave. & 39.27 & 44.49 & 49.19 & 50.02 & 51.81 & 59.50 & 60.79 & \underline{61.83} & 55.69 & \themebf{62.41} \\
\specialrule{\heavyrulewidth}{1pt}{0pt}
\end{tabularx}
\end{table*}

\begin{table*}[!t]
\centering
\caption{Comparison on single-task multimodal embedding benchmarks. Models are arranged as columns and sorted by Average in ascending order, with the two WePM variants listed last. The best result for each benchmark is highlighted and the second-best result is underlined; higher is better.}
\label{tab:single-task-embedding}
\setlength{\tabcolsep}{1.5pt}
\renewcommand{\arraystretch}{0.96}
\scriptsize
\begin{tabularx}{\textwidth}{@{}>{\raggedright\arraybackslash}m{0.24\textwidth}*{8}{>{\centering\arraybackslash}m{\dimexpr 0.0755\textwidth-3pt\relax}}>{\cellcolor{jcolour!30}\centering\arraybackslash}m{\dimexpr 0.0755\textwidth-3pt\relax}>{\cellcolor{jcolour!30}\centering\arraybackslash}m{\dimexpr 0.0755\textwidth-3pt\relax}@{}}
\specialrule{\heavyrulewidth}{0pt}{0pt}
\rowcolor{jcolour!15}Benchmark & \shortstack{Omni\\Embed\\Nemotron\\3B~\cite{nvidia2026omniembed}} & \shortstack{Jina\\Embedding\\v5-omni\\\cite{honicke2026jina}} & \shortstack{Embed-RL\\4B~\cite{jiang2026embedrl}} & \shortstack{UME-R1\\2B~\cite{lan2025umer1}} & \shortstack{Qwen3-VL\\Embedding\\2B~\cite{li2026qwen3vlembed}} & \shortstack{UEmbed\\4B~\cite{song2026uembed}} & \shortstack{WeMM\\Embedding\\2B~\cite{zhou2026wemm}} & \shortstack{Qwen3-VL\\Embedding\\8B~\cite{li2026qwen3vlembed}} & \cellcolor{jcolour!30}\themebf{\shortstack{WePM\\2B}} & \cellcolor{jcolour!30}\themebf{\shortstack{WePM\\4B}} \\
\specialrule{\lightrulewidth}{0pt}{0pt}
\rowcolor{gray!8} ImageNet-1K~\cite{deng2009imagenet} & 59.90 & 48.50 & 73.00 & 77.90 & 77.90 & 72.30 & \themebf{86.00} & \underline{82.60} & 78.90 & 81.60 \\
N24News~\cite{wang2022n24news} & 40.90 & 33.20 & 50.00 & \themebf{81.30} & 70.70 & 61.30 & \underline{80.60} & \underline{80.60} & 72.40 & 76.90 \\
\rowcolor{gray!8} VOC2007~\cite{everingham2010pascal} & 69.40 & 76.20 & 73.90 & 87.00 & 92.50 & 82.50 & \themebf{95.30} & \underline{93.70} & 89.10 & 90.40 \\
SUN397~\cite{xiao2010sun} & 63.00 & 61.80 & 66.20 & 80.10 & 81.50 & 75.00 & \themebf{83.20} & \underline{82.90} & 78.70 & 79.70 \\
\rowcolor{gray!8} Places365~\cite{zhou2017places} & 34.10 & 36.60 & 40.40 & 43.30 & 45.00 & 44.10 & \themebf{52.90} & \underline{47.90} & 44.20 & 44.40 \\
ImageNet-R~\cite{hendrycks2021imagenetr} & 84.10 & 84.20 & 83.70 & 90.20 & \underline{93.50} & 87.10 & 92.20 & \themebf{93.90} & 86.70 & 86.20 \\
\rowcolor{gray!8} DocVQA~\cite{mathew2021docvqa} & 8.90 & 21.20 & 68.80 & 93.40 & 94.90 & \underline{95.80} & 95.40 & \themebf{96.20} & 89.00 & 92.50 \\
ChartQA~\cite{masry2022chartqa} & 12.70 & 8.70 & 31.80 & 55.60 & 65.70 & 72.10 & \themebf{75.20} & \underline{74.30} & 65.30 & 72.50 \\
\rowcolor{gray!8} WebQA~\cite{chang2022webqa} & 47.80 & 38.00 & 40.90 & 84.50 & 90.10 & 90.20 & \underline{90.40} & \themebf{91.70} & 86.90 & 89.10 \\
MSCOCO~\cite{lin2014coco} & 21.80 & 25.90 & 70.00 & 66.60 & 81.40 & 81.40 & \underline{84.00} & \themebf{86.40} & 82.40 & 83.50 \\
\rowcolor{gray!8} RefCOCO~\cite{yu2016refcoco} & 29.70 & 29.90 & 76.00 & 82.30 & \themebf{96.10} & \underline{95.80} & 95.60 & 95.70 & 94.10 & 94.10 \\
RefCOCO-Matching~\cite{yu2016refcoco} & \themebf{100.00} & 60.50 & \themebf{100.00} & 83.60 & 82.80 & 91.50 & 80.50 & 90.90 & \underline{94.10} & 93.00 \\
\rowcolor{gray!8} ViDoSeek-doc~\cite{wang2025vidorag} & 88.95 & 88.83 & \themebf{89.29} & 78.74 & 84.60 & \underline{89.23} & 83.42 & 86.09 & 76.34 & 80.42 \\
CQADupstack-Unix~\cite{hoogeveen2015cqadupstack} & \themebf{51.17} & \underline{49.61} & 40.66 & 33.08 & 43.47 & 49.56 & 40.46 & 48.61 & 39.41 & 45.10 \\
\rowcolor{gray!8} FiQA-2018~\cite{maia2018fiqa} & \underline{52.91} & 49.17 & 35.81 & 18.36 & 42.30 & \themebf{53.45} & 50.33 & 46.00 & 45.24 & 51.17 \\
SCIDOCS~\cite{cohan2020specter} & 21.67 & \underline{23.14} & 21.45 & 7.98 & 21.89 & 22.51 & 16.04 & 22.91 & 21.91 & \themebf{24.32} \\
\rowcolor{gray!8} TREC-COVID~\cite{roberts2021treccovid} & 78.88 & \underline{78.95} & 52.29 & 39.91 & \themebf{85.93} & 76.08 & 38.66 & 66.39 & 63.37 & 65.24 \\
T2Retrieval~\cite{xiao2023cpack} & 81.98 & 83.18 & 66.59 & 54.65 & 81.40 & \underline{84.77} & 83.83 & \themebf{85.39} & 78.86 & 81.69 \\
\rowcolor{gray!8} MMarcoRetrieval~\cite{xiao2023cpack} & 77.34 & \themebf{82.39} & 67.29 & 60.01 & 76.63 & \underline{80.83} & 79.73 & 80.45 & 75.33 & 78.26 \\
CmedqaRetrieval~\cite{xiao2023cpack} & 32.83 & 39.16 & 31.15 & 25.29 & 37.15 & 38.67 & 38.11 & \themebf{41.39} & 37.54 & \underline{41.16} \\
\rowcolor{gray!8} SICK-R~\cite{marelli2014sick} & 73.19 & \themebf{91.62} & 78.16 & 76.70 & 81.72 & \underline{83.28} & 80.92 & 80.93 & 78.06 & 78.98 \\
STS22~\cite{chen2022sts22} & 67.53 & \themebf{71.74} & \underline{70.89} & 52.90 & 70.34 & 69.16 & 66.49 & 70.66 & 65.06 & 70.15 \\
\rowcolor{gray!8} STSBenchmark~\cite{cer2017stsbenchmark} & 81.03 & \themebf{94.82} & 85.36 & 79.73 & 84.39 & \underline{88.34} & 81.20 & 84.79 & 79.56 & 81.50 \\
AmazonCounterfactual~\cite{oneill2021counterfactual} & 75.72 & \themebf{93.15} & 68.75 & 63.66 & 69.15 & 81.75 & \underline{83.43} & 72.24 & 74.03 & 73.28 \\
\rowcolor{gray!8} Banking77~\cite{casanueva2020banking77} & 83.00 & 91.46 & 80.54 & 75.40 & 81.23 & 84.36 & 88.83 & 84.63 & \underline{92.34} & \themebf{92.56} \\
MTOPDomain~\cite{li2021mtop} & 95.67 & \themebf{98.93} & 93.23 & 92.66 & 94.71 & \underline{96.71} & 95.96 & 95.51 & 94.41 & 94.80 \\
\rowcolor{gray!8} ToxicConversations~\cite{jigsaw2019toxicity} & 66.79 & \themebf{93.64} & 67.70 & 67.39 & 64.65 & 72.42 & 86.36 & 67.38 & 86.31 & \underline{86.39} \\
BiorxivClusteringP2P~\cite{muennighoff2022mteb} & 38.42 & \underline{47.93} & 39.59 & 30.08 & 40.83 & 43.76 & \themebf{50.28} & 39.03 & 37.02 & 36.12 \\
\rowcolor{gray!8} TwentyNewsgroups~\cite{lang1995newsweeder} & 29.95 & \underline{54.45} & 51.23 & 48.05 & 51.57 & 52.96 & 53.26 & 54.20 & 52.83 & \themebf{56.59} \\
AskUbuntuDupQuestions~\cite{lei2016askubuntu,wang2021tsdae} & 22.31 & 43.44 & 33.63 & 27.94 & 37.80 & 42.60 & 32.86 & 41.43 & \underline{62.34} & \themebf{64.29} \\
\specialrule{\lightrulewidth}{0pt}{0pt}
Ave. & 56.39 & 60.01 & 61.61 & 61.94 & 70.73 & 71.98 & 72.05 & \underline{72.83} & 70.73 & \themebf{72.86} \\
\specialrule{\heavyrulewidth}{1pt}{0pt}
\end{tabularx}

\vspace{4pt}
\caption{Comparison on single-task decision benchmarks. The best result in each column is highlighted and the second-best result is underlined; higher is better. Models are sorted by Average in ascending order, with the two WePM variants listed last.}
\label{tab:single-task-decision}
\setlength{\tabcolsep}{2.4pt}
\renewcommand{\arraystretch}{1.08}
\scriptsize
\begin{tabularx}{\textwidth}{@{}>{\raggedright\arraybackslash}X>{\centering\arraybackslash}m{0.055\textwidth}>{\centering\arraybackslash}m{0.070\textwidth}>{\centering\arraybackslash}m{0.070\textwidth}>{\centering\arraybackslash}m{0.085\textwidth}>{\centering\arraybackslash}m{0.120\textwidth}>{\centering\arraybackslash}m{0.110\textwidth}>{\centering\arraybackslash}m{0.105\textwidth}!{\vrule width \lightrulewidth}>{\centering\arraybackslash}m{0.055\textwidth}@{}}
\specialrule{\heavyrulewidth}{0pt}{0pt}
\rowcolor{jcolour!15}\rule[-1pt]{0pt}{16pt}Model & \shortstack{S1MB\\\cite{hotchpotch2026s1mb}} & \shortstack{JevBench\\\cite{jevbench2026}} & \shortstack{Open-Jev\\\cite{caiz2026openjev}} & \shortstack{JevAny (MM)\\\cite{simplejev2026jevany}} & \shortstack{Valen-VDB (Image)\\\cite{valenteam2026vdb}} & \shortstack{Valen-General5k\\\cite{valenteam2026general5k}} & \shortstack{DecisionBench\\\cite{hannolabs2026decisionbench}} & Ave. \\
\specialrule{\lightrulewidth}{0pt}{0pt}
\rowcolor{gray!8} SmolVLM2-2B~\cite{huggingface2025smolvlm2} & 6.36 & 42.61 & 52.17 & 3.27 & 40.55 & 33.36 & 11.93 & 27.18 \\
OmniJev-2B~\cite{tinnel2026omnijev} & 12.09 & 38.41 & 71.83 & 8.73 & 46.86 & 32.55 & 30.13 & 34.37 \\
\rowcolor{gray!8} JEV-9B~\cite{autotrust2026jev9b} & 27.73 & 53.83 & 87.22 & 12.84 & 24.19 & 16.69 & 23.36 & 35.12 \\
Phi-3.5-vision-4B~\cite{microsoft2024phi35vision} & 19.86 & 54.29 & 54.88 & 1.54 & 48.59 & 36.00 & 33.00 & 35.45 \\
\rowcolor{gray!8} Gemma-4-E2B-it~\cite{gemmateam2026gemma4} & 20.22 & 55.08 & 65.36 & 2.78 & 39.80 & 29.17 & 37.06 & 35.64 \\
MiniCPM-V-4.0~\cite{openbmb2025minicpmv4} & 15.58 & 48.98 & 61.98 & 2.13 & 48.62 & 36.00 & 37.96 & 35.89 \\
\rowcolor{gray!8} InternVL3-8B~\cite{zhu2025internvl3} & 16.91 & 54.71 & 69.60 & 2.96 & 46.66 & 36.07 & 34.95 & 37.41 \\
Gemma-4-E4B-it~\cite{gemmateam2026gemma4} & 30.28 & 58.75 & 73.73 & 3.56 & 39.85 & 32.17 & 39.04 & 39.62 \\
\rowcolor{gray!8} Qwen3.5-4B~\cite{qwen3.5} & 27.86 & 59.14 & 74.50 & 5.79 & 50.43 & 37.26 & 26.35 & 40.19 \\
MiniCPM-o-4.5-9B~\cite{openbmb2026minicpmo45} & 30.76 & 57.84 & 78.35 & 3.97 & 50.32 & 38.97 & 49.34 & 44.22 \\
\rowcolor{gray!8} OmniJev-4B~\cite{tinnel2026omnijev} & 28.64 & 48.98 & 81.01 & 18.48 & 52.11 & 45.41 & 35.27 & 44.27 \\
Qwen3.5-9B~\cite{qwen3.5} & 32.19 & 60.32 & 77.11 & 4.08 & 54.69 & 54.77 & 27.50 & 44.38 \\
\rowcolor{gray!8} d3-lite~\cite{vllmsr2026d3} & 35.61 & 57.46 & 85.00 & 14.55 & 42.28 & 37.78 & 41.40 & 44.87 \\
StartLux-Decision-v1-0.8B~\cite{startlux2026decision} & 43.14 & 55.94 & 85.95 & 14.23 & 43.40 & 34.20 & 40.90 & 45.39 \\
\rowcolor{gray!8} d1-3B~\cite{liquidai2026d1} & 35.02 & 59.86 & 83.98 & 15.55 & 51.59 & 40.88 & 45.31 & 47.46 \\
d3-nano~\cite{vllmsr2026d3} & 43.27 & 61.12 & 86.89 & 20.15 & 48.56 & 54.56 & 49.70 & 52.04 \\
\rowcolor{gray!8} NeoHorse-Jev-4B~\cite{tokenrhythm2026neohorse} & 39.14 & 59.60 & 88.04 & 17.84 & 53.99 & 60.52 & 46.09 & 52.17 \\
Clef-Flash~\cite{cloudflare2026clef} & 42.82 & 59.97 & 86.56 & 21.19 & \underline{59.18} & 54.08 & 47.78 & 53.08 \\
\rowcolor{gray!8} StartLux-Decision-v1-4B~\cite{startlux2026decision} & \themebf{55.94} & \underline{62.39} & 94.14 & 24.41 & \themebf{62.45} & 66.47 & \underline{51.28} & 59.58 \\
d3-mini-4.5B~\cite{vllmsr2026d3} & \underline{55.78} & \themebf{63.49} & \themebf{97.04} & \underline{25.17} & 57.32 & 74.11 & \themebf{54.50} & \underline{61.06} \\
\rowcolor{jcolour!30}\themebf{WePM-2B} & 41.64 & 58.74 & 92.64 & 15.99 & 56.47 & \underline{81.51} & 43.68 & 55.81 \\
\rowcolor{jcolour!30}\themebf{WePM-4B} & 54.04 & 60.91 & \underline{95.74} & \themebf{25.60} & 58.36 & \themebf{82.99} & 50.13 & \themebf{61.11} \\
\specialrule{\heavyrulewidth}{1pt}{0pt}
\end{tabularx}
\end{table*}

\subsection{Comparison with Unified Methods on Shared Capabilities}
\label{sec:sota-comparison}

Existing unified baselines provide multimodal retrieval and generation but no native typed-decision path. We therefore compare the complete WePM model with CAFe~\cite{yu2025cafe}, FROMAGe~\cite{koh2023fromage}, MMGen~\cite{wang2025mmgen}, and ULME~\cite{liu2026ulme} on the two capability families shared by all methods. Table~\ref{tab:sota-comparison} summarizes performance across the five evaluation groups, together with the shared-capability Joint score defined in Section~\ref{sec:experimental-details}.

\par\Needspace*{0.30\textheight}
\begin{wraptable}{r}{0.64\linewidth}
\centering
\caption{Comparison with state-of-the-art unified methods on the shared retrieval--generation capabilities. The best result in each column is shown in bold. Joint is the two-level arithmetic mean defined in Section~\ref{sec:experimental-details}; higher is better.}
\label{tab:sota-comparison}
\setlength{\tabcolsep}{2pt}
\renewcommand{\arraystretch}{1.12}
\begin{tabularx}{\linewidth}{@{}>{\raggedright\arraybackslash}p{0.24\linewidth}*{6}{>{\centering\arraybackslash}X}@{}}
\specialrule{\heavyrulewidth}{0pt}{0pt}
\rowcolor{jcolour!15} & \multicolumn{2}{c}{Embedding} & \multicolumn{3}{c}{Generation} & \\
\rowcolor{jcolour!15}\multirow{-2}{*}{Method} & ImageRet. & TextRet. & Math & VQA & OCR & \multirow{-2}{*}{Joint} \\
\specialrule{\lightrulewidth}{0pt}{0pt}
\rowcolor{gray!8}CAFe~\cite{yu2025cafe} & 22.83 & 30.30 & 7.18 & 13.71 & 40.97 & 23.59 \\
FROMAGe~\cite{koh2023fromage} & 32.30 & 33.12 & 22.60 & 34.29 & 61.86 & 36.15 \\
\rowcolor{gray!8}MMGen~\cite{wang2025mmgen} & 30.93 & 33.26 & 24.17 & 39.05 & 60.11 & 36.60 \\
ULME~\cite{liu2026ulme} & 34.13 & 32.78 & 28.16 & 41.46 & 58.93 & 38.15 \\
\rowcolor{jcolour!30}\themebf{WePM} & \themebf{41.30} & \themebf{35.90} & \themebf{58.60} & \themebf{60.84} & \themebf{81.18} & \themebf{52.74} \\
\specialrule{\heavyrulewidth}{1pt}{0pt}
\end{tabularx}
\end{wraptable}

\themebf{Overall performance.} WePM achieves the best result in all five evaluation groups. Relative to the strongest baseline in each group, it improves ImageRet. and TextRet. by 7.17 and 2.64 points, respectively, while yielding substantially larger gains of 30.44 points on Math, 19.38 points on VQA, and 19.32 points on OCR. The consistent improvements across both task families demonstrate that the gain in generation capability does not come at the expense of retrieval quality.

\themebf{Balance between retrieval and generation.} WePM obtains aggregate Embedding and Generation scores of 38.60 and 66.87, respectively, surpassing the strongest baseline aggregates of 33.46 and 42.85, both achieved by ULME. Consequently, WePM reaches a Joint score of 52.74, outperforming ULME by 14.59 points. The larger margin on generation, together with the simultaneous improvement in retrieval, indicates a stronger overall balance between the two capabilities. Section~\ref{sec:ablation-studies} further attributes this improvement to the visual-token allocation, distillation, and multitask-optimization components.

\finishwrapfigure

\subsection{Ablation Studies}
\label{sec:ablation-studies}

We conduct controlled ablations on the shared representation--generation protocol along three dimensions: the visual-token budget in MGVB, the generation-side distillation strategy in ULKD, and the task-weighting and execution strategies in SEMO. This common two-capability setting provides dense benchmark coverage and keeps the evaluation consistent across all variants; complete three-capability behavior is assessed separately under the full-capability protocol.

\subsubsection{MGVB: Visual-Token Budget Ablation}

\par\Needspace*{0.30\textheight}
\begin{wraptable}{r}{0.64\linewidth}
\centering
\caption{Effect of the MGVB visual-token budget. Joint is the two-level arithmetic mean defined in Section~\ref{sec:experimental-details}; higher is better.}
\label{tab:mgvb-budget}
\setlength{\tabcolsep}{2pt}
\renewcommand{\arraystretch}{1.08}
\begin{tabularx}{\linewidth}{@{}>{\raggedright\arraybackslash}p{0.17\linewidth}>{\centering\arraybackslash}p{0.12\linewidth}*{6}{>{\centering\arraybackslash}X}@{}}
\specialrule{\heavyrulewidth}{0pt}{0pt}
\rowcolor{jcolour!15} & & \multicolumn{2}{c}{Embedding} & \multicolumn{3}{c}{Generation} & \\
\rowcolor{jcolour!15}\multirow{-2}{*}{Method} & \multirow{-2}{*}{$B$} & ImageRet. & TextRet. & Math & VQA & OCR & \multirow{-2}{*}{Joint} \\
\specialrule{\lightrulewidth}{0pt}{0pt}
\rowcolor{gray!8}Baseline & -- & 30.27 & 33.06 & 26.43 & 40.58 & 64.59 & 37.77 \\
MGVB & 128 & 34.13 & \themebf{35.16} & 30.12 & 44.90 & 58.53 & 39.58 \\
\rowcolor{gray!8}MGVB & 256 & \themebf{37.77} & 34.60 & \themebf{30.94} & 44.27 & 65.55 & 41.55 \\
\rowcolor{jcolour!30}\themebf{MGVB} & \themebf{512} & 36.97 & 34.88 & 30.90 & \themebf{45.52} & \themebf{67.21} & \themebf{41.90} \\
\specialrule{\heavyrulewidth}{1pt}{0pt}
\end{tabularx}
\end{wraptable}

To assess the effect of visual capacity, we compare the baseline pathway, which preserves the full uniformly organized visual-token sequence, with MGVB under output budgets of 128, 256, and 512 tokens. All variants are evaluated on the same five groups using the Joint metric defined in Section~\ref{sec:experimental-details}. The baseline, which is shared with the subsequent ULKD and SEMO ablations, obtains a Joint score of 37.77.

\themebf{Effectiveness under different token budgets.} With only 128 tokens, MGVB improves Joint by 1.81 points, from 37.77 to 39.58, and outperforms the baseline in four of the five groups. OCR decreases by 6.06 points, suggesting that aggressive compression preferentially removes localized text evidence. Increasing the budget to 256 yields improvements in all five groups and raises Joint to 41.55, including a 7.50-point gain in ImageRet. and a recovery of OCR from 58.53 to 65.55. At $B=512$, MGVB achieves the highest Joint score of 41.90, exceeding the baseline by 4.13 points, and obtains the strongest VQA and OCR results among the tested budgets. These results show that multi-granularity allocation becomes increasingly effective as sufficient capacity is available to preserve fine-grained evidence.

\themebf{Diminishing returns and capability trade-off.} Increasing $B$ from 128 to 256 improves Joint by 1.97 points, whereas the gain from 256 to 512 is only 0.35 points. Over the latter interval, the aggregate Embedding score decreases slightly from 36.19 to 35.93, while the Generation score increases from 46.92 to 47.88, driven primarily by gains of 1.25 points in VQA and 1.66 points in OCR. We therefore adopt $B=512$ as the operating point: it provides the best overall result while avoiding further consumption of the context window shared by visual and text tokens. The observed trend also exposes a scale-dependent balance between retrieval semantics and local visual detail; isolating the effects of the individual token banks and the query-aware selector remains beyond what this budget-only ablation can establish.

\finishwrapfigure

\par\Needspace*{0.27\textheight}
\begin{wrapfigure}{r}{0.55\linewidth}
\centering
\includegraphics[width=\linewidth]{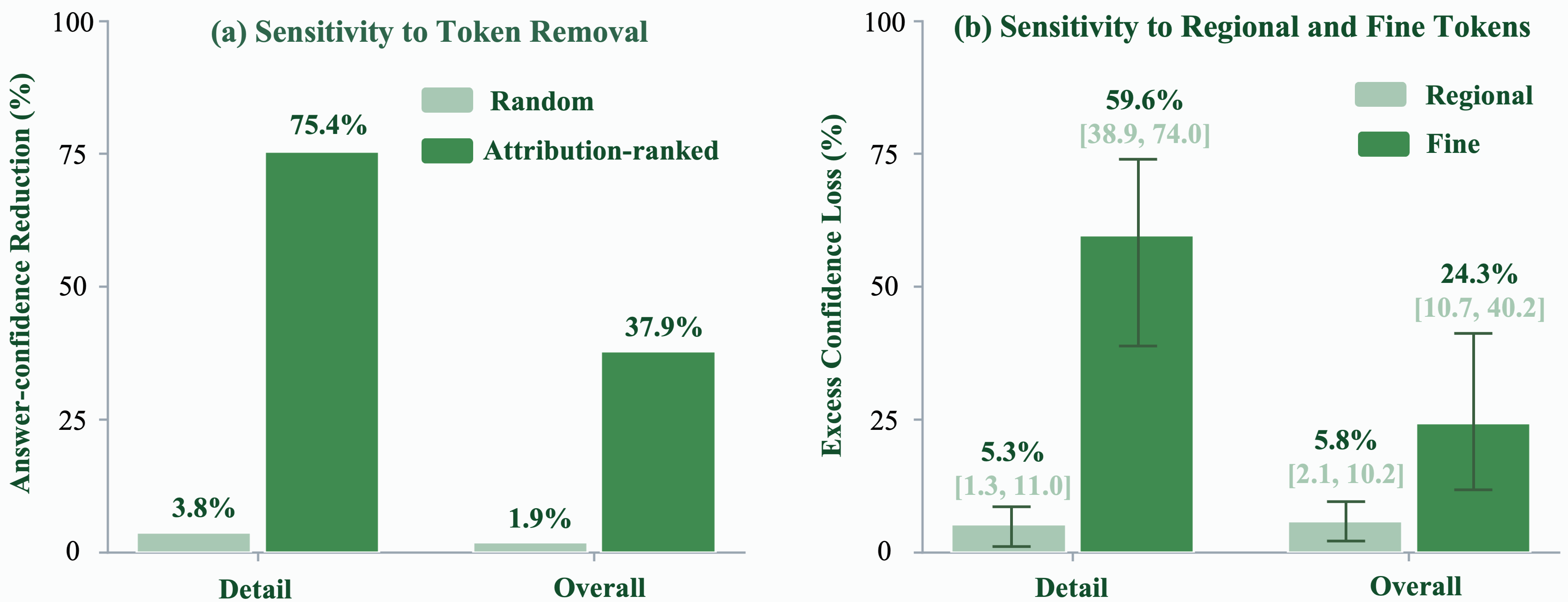}
\caption{Visual-evidence removal in MGVB. (a) Matched random and attribution-ranked masking. (b) Equal-count regional and fine masking; error bars denote 95\% bootstrap confidence intervals.}
\label{fig:mgvb-evidence-removal}
\end{wrapfigure}

\themebf{Sensitivity to Visual Evidence Removal.} The budget ablation characterizes the effect of visual-token capacity, but does not reveal whether the retained tokens contribute uniformly or whether evidence at different granularities is interchangeable. We therefore conduct a controlled masking analysis across multiple chart-image collections with $B=256$. For each image, the MGVB visual bank is captured once and fixed throughout the analysis. Regional and fine token groups are ranked according to the reduction in reference-answer log probability caused by masking their visual content under the original question, and their effects are evaluated on a paraphrased question. As shown in Fig.~\ref{fig:mgvb-evidence-removal}(a), randomly masking 16 regional and 16 fine tokens reduces answer confidence by only 3.8\% for detail questions and 1.9\% for overall questions, whereas masking the same number of attribution-ranked tokens causes reductions of 75.4\% and 37.9\%, respectively. Figure~\ref{fig:mgvb-evidence-removal}(b) further isolates each granularity through equal-count, single-bank masking. Relative to matched random masks, fine-token removal causes substantially larger confidence reductions than regional-token removal for both detail questions (59.6\% versus 5.3\%) and overall questions (24.3\% versus 5.8\%). These results reveal an asymmetric organization of visual evidence in MGVB: precise local evidence is concentrated in a small set of influential fine tokens and is difficult to recover once removed, whereas broader visual information is distributed more redundantly and is therefore more tolerant to incidental token loss. This asymmetry suggests that visual-token compression should prioritize evidence-aware allocation over uniform reduction, preserving fine-grained capacity while exploiting redundancy in more replaceable representations.

\finishwrapfigure

\subsubsection{ULKD: Dual-Teacher Distillation on Partially Paired Data}
\label{sec:ulkd-ablation}

We evaluate two aspects of ULKD: the benefit of introducing capability-specific teachers on partially paired data and the choice of generation-side distillation strategy. Both teacher-augmented variants use the same frozen Qwen3.5-27B generation teacher, embedding teacher, prompt pool, and original hard objectives. They differ only in generation supervision. Data distillation trains on teacher-generated sequences as fixed pseudo-targets, whereas on-policy distillation (OPD) samples student trajectories and matches the teacher's next-token distributions on the visited prefixes. This controlled comparison isolates the effect of sequence-level pseudo-targets versus on-policy distribution matching. Table~\ref{tab:ulkd-distillation} reports the evaluation groups and Joint metric defined in Section~\ref{sec:experimental-details}.

\par\Needspace*{0.30\textheight}
\begin{wraptable}{r}{0.64\linewidth}
\centering
\caption{ULKD ablation on teacher supervision and generation-side distillation. All teacher-augmented variants use the same embedding teacher; higher is better.}
\label{tab:ulkd-distillation}
\setlength{\tabcolsep}{2pt}
\renewcommand{\arraystretch}{1.08}
\begin{tabularx}{\linewidth}{@{}>{\raggedright\arraybackslash}p{0.27\linewidth}*{6}{>{\centering\arraybackslash}X}@{}}
\specialrule{\heavyrulewidth}{0pt}{0pt}
\rowcolor{jcolour!15} & \multicolumn{2}{c}{Embedding} & \multicolumn{3}{c}{Generation} & \\
\rowcolor{jcolour!15}\multirow{-2}{*}{Method} & ImageRet. & TextRet. & Math & VQA & OCR & \multirow{-2}{*}{Joint} \\
\specialrule{\lightrulewidth}{0pt}{0pt}
\rowcolor{gray!8}Baseline & 30.27 & 33.06 & 26.43 & 40.58 & 64.59 & 37.77 \\
Data KD + Emb. teacher & \themebf{42.50} & 32.58 & 42.07 & \themebf{62.05} & 79.24 & 49.33 \\
\rowcolor{jcolour!30}\themebf{OPD + Emb. teacher} & 41.47 & \themebf{33.59} & \themebf{50.66} & 61.82 & \themebf{85.13} & \themebf{51.70} \\
\specialrule{\heavyrulewidth}{1pt}{0pt}
\end{tabularx}
\end{wraptable}

\themebf{Benefit of teacher supervision.} The data-distillation configuration with the embedding teacher raises Joint from 37.77 to 49.33, an improvement of 11.56 points over the hard-label baseline. Replacing generation-side data distillation with OPD further increases Joint to 51.70. The resulting dual-teacher configuration outperforms the baseline by 13.93 points and improves all five evaluation groups, demonstrating that partially paired examples can provide effective cross-capability supervision without additional paired annotations.

\themebf{Advantage of on-policy distribution matching.} Under the same embedding teacher, OPD improves Joint by 2.37 points over data distillation. The gain is concentrated on generation: Math and OCR increase by 8.59 and 5.89 points, respectively, while VQA changes marginally by $-0.23$, raising the aggregate Generation score from 61.12 to 65.87. Meanwhile, the Embedding score remains effectively unchanged (37.54 versus 37.53), with a 1.03-point decrease in ImageRet. offset by a 1.01-point increase in TextRet. These results support supervising student-visited prefixes with the teacher's full token distribution rather than committing open-ended prompts to fixed sequence-level pseudo-targets.

\finishwrapfigure

\par\Needspace*{0.30\textheight}
\begin{wrapfigure}{r}{0.60\linewidth}
\centering
\includegraphics[width=\linewidth]{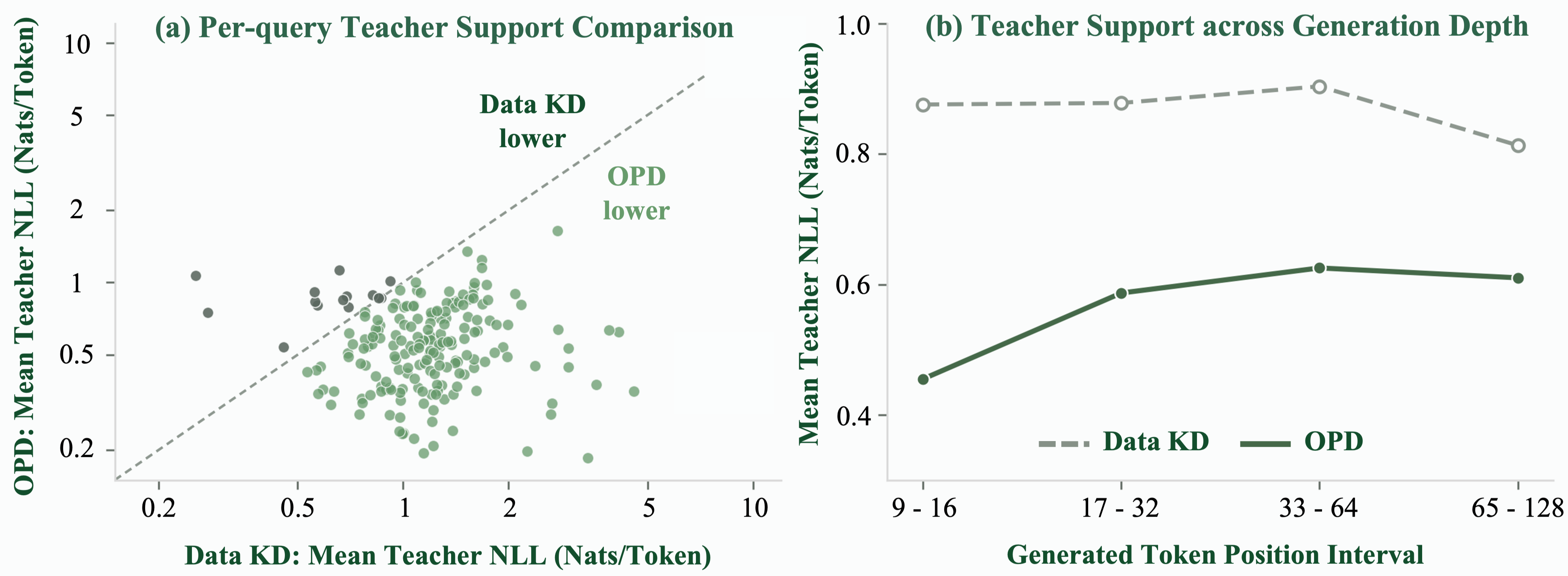}
\caption{Teacher support under free-running generation. (a) Per-query mean teacher NLL of continuations independently generated by the historical Data KD and OPD checkpoints on 195 open-ended retrieval queries; points below the diagonal indicate lower NLL, and thus stronger teacher support, for OPD. (b) Mean teacher NLL across token-position intervals of the student-generated continuation, excluding the first eight positions. An interval includes only queries for which both students generated tokens reaching that position, so the number of paired queries decreases with generation depth. Lower NLL indicates stronger teacher support.}
\label{fig:ulkd-teacher-support}
\end{wrapfigure}

\themebf{Teacher support under free-running generation.} The comparison above shows that OPD improves downstream generation, but it does not establish whether on-policy supervision keeps the student within regions that the teacher supports. We therefore let the two historical checkpoints generate freely on 195 held-out open-ended retrieval queries, for which no unique continuation exists, and have the frozen Qwen3.5-27B teacher score every realized token conditioned only on the query and the preceding student-generated prefix; no complete response is designated as the target. As shown in Fig.~\ref{fig:ulkd-teacher-support}(a), OPD yields a lower mean teacher NLL than Data KD (0.584 versus 1.323 nats/token), a paired reduction of 0.739 (95\% bootstrap CI [0.618, 0.892]), and is lower on 181 of 195 queries. The advantage is not confined to the opening: after excluding positions 1--8, where the gap is largest, OPD remains lower in every subsequent interval (Fig.~\ref{fig:ulkd-teacher-support}(b)), as well as on the matched first 32 tokens (1.712 versus 0.519) and on the 88 queries whose outputs are both CJK-free (0.543, 95\% CI [0.347, 0.820]). These results are consistent with sequence-level pseudo-targets transferring less reliably to free-running generation than supervision on student-visited prefixes. 
Together, these results indicate that on-policy supervision on student-visited prefixes produces continuations that remain more consistently within the teacher's high-probability region than sequence-level pseudo-targets.

\finishwrapfigure

\subsubsection{SEMO: Fixed Weighting and Sequential Joint Updates}

We evaluate SEMO on the controlled representation--generation slice for optimization stability and joint-update memory feasibility. All variants use the same exposure schedule---three embedding and five generation batches per logical update---and sequential execution. Standard FAMO measures task progress as $q_{i,t}=\log D_{i,t}^{\mathrm{pre}}-\log D_{i,t}^{\mathrm{post}}$, where $D_{i,t}^{\mathrm{pre}}$ and $D_{i,t}^{\mathrm{post}}$ are the transformed losses before and after update $t$; this feedback affects the next logical update. We test two diagnostics: \emph{FAMO w/o delay} forms weights from the current pre-update losses rather than the preceding post-update state, while \emph{FAMO w/o normalization} uses the raw decrease $q_{i,t}^{\mathrm{raw}}=D_{i,t}^{\mathrm{pre}}-D_{i,t}^{\mathrm{post}}$. We compare these variants with fixed SEMO ratios and the joint-training baseline.

\par\Needspace*{0.14\textheight}
\noindent
\begin{minipage}[t]{0.28\linewidth}
\vspace{0pt}
\themebf{Fixed coefficients yield the strongest joint performance.} All SEMO ratios outperform the adaptive variants (49.15--49.78 versus 45.72--47.99 Joint). Equal weighting reaches 49.78, improving over standard FAMO by 3.55 points and the joint-training baseline by 12.01 points. Its Embedding/Generation scores of 41.18/58.39 versus FAMO's 35.11/57.36 show stronger retrieval without sacrificing generation.
\end{minipage}\hfill
\begin{minipage}[t]{0.70\linewidth}
\vspace{0pt}
\centering
\captionof{table}{SEMO and FAMO under the same task-exposure schedule; higher Joint is better.}
\label{tab:semo-weighting}
\setlength{\tabcolsep}{2pt}
\renewcommand{\arraystretch}{0.96}
\begin{tabularx}{\linewidth}{@{}>{\raggedright\arraybackslash}p{0.25\linewidth}>{\centering\arraybackslash}p{0.16\linewidth}*{6}{>{\centering\arraybackslash}X}@{}}
\specialrule{\heavyrulewidth}{0pt}{0pt}
\rowcolor{jcolour!15} & & \multicolumn{2}{c}{Embedding} & \multicolumn{3}{c}{Generation} & \\
\rowcolor{jcolour!15}\multirow{-2}{*}{Method} & \multirow{-2}{*}{\shortstack{Weight\\$\lambda_{\mathrm{emb}}/\lambda_{\mathrm{gen}}$}} & ImageRet. & TextRet. & Math & VQA & OCR & \multirow{-2}{*}{Joint} \\
\specialrule{\lightrulewidth}{0pt}{0pt}
\rowcolor{gray!8}Baseline & -- & 30.27 & 33.06 & 26.43 & 40.58 & 64.59 & 37.77 \\
FAMO (standard) & dynamic & 34.00 & 36.21 & 38.29 & 55.08 & \themebf{78.71} & 46.23 \\
\rowcolor{gray!8}FAMO w/o delay & dynamic & 36.20 & 35.81 & 37.58 & 51.66 & 77.07 & 45.72 \\
FAMO w/o norm. & dynamic & 40.33 & 36.69 & 40.48 & 54.09 & 77.82 & 47.99 \\
\rowcolor{gray!8}SEMO & 0.3/0.7 & \themebf{44.67} & \themebf{38.13} & 36.84 & \themebf{57.94} & 77.90 & 49.48 \\
\rowcolor{jcolour!30}\themebf{SEMO} & \themebf{0.5/0.5} & 44.57 & 37.78 & \themebf{42.40} & 55.02 & 77.76 & \themebf{49.78} \\
\rowcolor{gray!8}SEMO & 0.7/0.3 & 41.53 & 37.64 & 41.86 & 56.14 & 78.17 & 49.15 \\
\specialrule{\heavyrulewidth}{1pt}{0pt}
\end{tabularx}
\end{minipage}
\par\addvspace{\intextsep}

\themebf{Fixed weighting is robust to the coefficient ratio.} Across the three tested ratios, Joint varies by only 0.63 points. Assigning a larger coefficient to generation ($0.3/0.7$) produces the best Embedding score of 41.40, whereas emphasizing embedding ($0.7/0.3$) produces the best Generation score of 58.72. Equal weighting remains within 0.22 points of the best Embedding score and 0.33 points of the best Generation score, yielding the strongest combined result. This non-monotonic response indicates that loss coefficients interact through the shared representation rather than independently controlling their corresponding downstream capabilities.

\themebf{FAMO is sensitive to its feedback definition.} Removing the one-update delay increases Embedding from 35.11 to 36.01 but decreases Generation from 57.36 to 55.44, reducing Joint from 46.23 to 45.72. Modifying the feedback normalization is more effective, raising Joint to 47.99, yet it still underperforms every fixed-weight configuration. These results rule out feedback staleness as the sole explanation and suggest that short-term loss progress is itself an unstable signal for balancing the two capabilities in this setting.

\themebf{Sequential accumulation substantially reduces peak memory.} We profile ten logical updates on eight 96-GiB GPUs with TP2/DP4, equal task coefficients, and three embedding plus five generation batches per update. Sequential execution uses $74.42\pm5.45$ GiB on average and peaks at 76.23 GiB, whereas concurrent execution reaches 209.98 GiB, approximately $2.8\times$ the sequential footprint, and exceeds the 96-GiB device capacity in all ten updates. The measured activation increment remains stable for each batch (approximately 22.4 GiB for embedding and 20.7 GiB for generation) and is released after backpropagation rather than accumulating across the update. SEMO therefore retains only the active batch graph while forming one joint optimizer update; fixed coefficients additionally avoid the post-update loss re-evaluation required by standard FAMO. As an implementation check under distributed bf16 training, its accumulated gradient matches a joint-backward reference across all module groups (cosine similarity $\geq 0.99995$), confirming that the memory reduction does not introduce a measurable directional bias.

\vspace{-10pt}
\subsection{Heterogeneous Capability Density}

\begin{wrapfigure}{r}{0.45\linewidth}
\centering
\includegraphics[width=\linewidth]{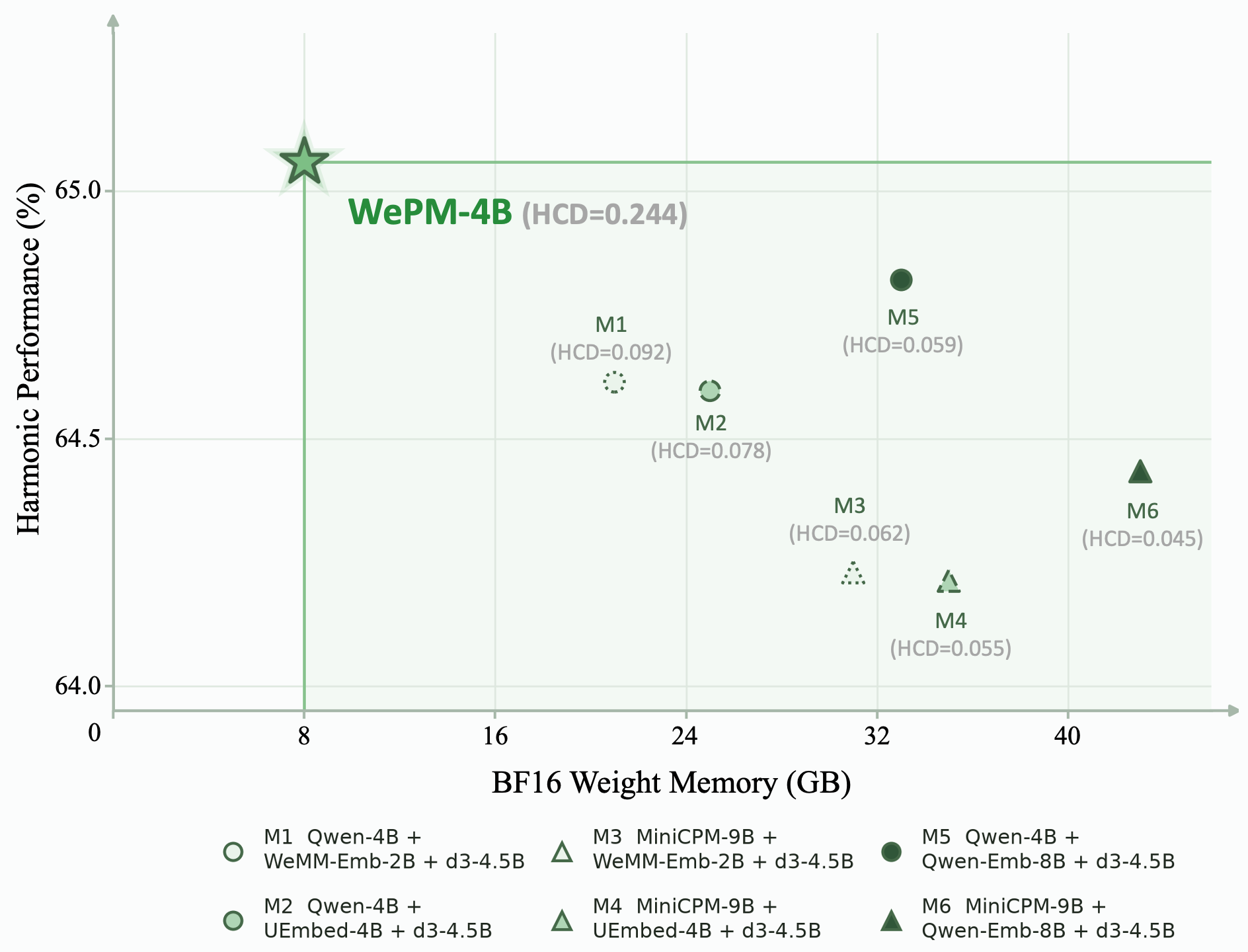}
\caption{Heterogeneous capability density under BF16 resident weight memory. WePM jointly supports generation, representation, and decision with one shared parameter set, while M1--M6 combine separate specialist models. Harmonic performance is the equal-weight harmonic mean across the three capability families.}
\label{fig:hcd-memory-quality}
\end{wrapfigure}

\themebf{Comparison protocol.} We construct six representative separate-model stacks by taking the Cartesian product of two strong generation specialists (Qwen3.5-4B and MiniCPM-o-4.5-9B) and three embedding specialists spanning 2B, 4B, and 8B scales (WeMM-Embedding-2B, UEmbed-4B, and Qwen3-VL-Embedding-8B). Every stack uses d3-mini-4.5B as its decision component because it is the strongest separate decision specialist in our evaluation. This design varies generation and embedding capacity while holding decision quality fixed, and covers compact, intermediate, and high-memory deployment choices rather than selecting a favorable baseline. M1 and M3 pair the two generators with the 2B embedding model, M2 and M4 use the 4B model, and M5 and M6 use the 8B model, as detailed in the legend of Fig.~\ref{fig:hcd-memory-quality}.

\themebf{Metric interpretation.} For each separate stack, resident memory is the sum of the three component models' BF16 weights, using two bytes per parameter; WePM counts its shared 4B parameters only once because all three capabilities are exposed by the same checkpoint. We first summarize balanced task quality with the equal-weight harmonic mean of the normalized Generation, Embedding, and Decision scores. The harmonic mean penalizes a stack whose apparent strength is driven by only one capability. With all three families active ($K=3$), HCD further divides this balanced quality by resident memory, so moving upward indicates stronger balanced performance, moving left indicates a smaller deployment footprint, and a larger HCD indicates that more heterogeneous capability is delivered per unit of memory.

\themebf{Results.} As shown in Fig.~\ref{fig:hcd-memory-quality}, the six separate-model stacks require 21--43~GB of BF16 weight memory and obtain harmonic performance of 64.21--64.82. Increasing specialist capacity therefore raises memory substantially without surpassing WePM in balanced quality. WePM supports all three families with one 8-GB parameter set and reaches a harmonic score of 65.06, placing it above and to the left of every separate stack. Its HCD is 0.244, compared with 0.092 for M1, the strongest separate stack by HCD, yielding a $2.64\times$ improvement. Against M1, WePM reduces resident weight memory by 61.9\% (8 versus 21~GB) while improving harmonic performance by 0.45 points. Even M5, the highest-quality separate stack, reaches only 64.82 at 33~GB. These results show that WePM's advantage comes not merely from using fewer parameters, but from sharing them across capabilities while preserving stronger and more balanced task quality.

\finishwrapfigure

\vspace{-25pt}
\section{Positioning Against Existing Approaches}
\label{sec:related-work}

\subsection{Unified Representation--Generation Models and Distillation}

Contrastive vision--language pretraining, exemplified by CLIP~\cite{radford2021clip}, established transferable representations for image--text matching. Recent work has extended generative vision--language backbones into general-purpose embedding models. VLM2Vec learns instruction-conditioned multimodal embeddings~\cite{jiang2025vlm2vec}, while UniME combines knowledge distillation with hard-negative training for universal multimodal retrieval~\cite{gu2025unime}. A complementary line of research jointly supports retrieval and generation within one model. FROMAGe augments a frozen generative language model with image retrieval~\cite{koh2023fromage}; GRIT combines embedding and generative instruction tuning~\cite{muennighoff2025grit}; CAFe couples contrastive and autoregressive objectives~\cite{yu2025cafe}; and MMGen trains a unified multimodal model for both embedding and generation~\cite{wang2025mmgen}. GEM further connects generated query analysis with dense retrieval, although its scope is limited to text retrieval~\cite{shen2026gem}. Retrieval and generation have also developed distinct forms of teacher supervision. RocketQAv2 performs dynamic listwise distillation between a dense retriever and a re-ranker~\cite{ren2021rocketqav2}, while BiXSE learns from probabilistic graded-relevance targets for individual query--document pairs~\cite{tsirigotis2025bixse}. For language generation, GKD trains on student-generated sequences to reduce exposure mismatch between fixed training targets and student-visited states~\cite{agarwal2024gkd}; subsequent on-policy distillation similarly evaluates continuations sampled from the student~\cite{thinkingmachines2025opd}. Cross-world on-policy distillation extends this principle to VLMs by supervising prediction changes under interventions to answer-relevant visual evidence~\cite{sun2026crossworld}.

WePM follows the shared-backbone paradigm but targets heterogeneous capability under compact-model constraints. Beyond coupling indexable embeddings with open-ended generation, it introduces schema-constrained typed decision within the same backbone. Its ULKD component brings retrieval and generation distillation into a common distribution-matching framework without conflating their prediction spaces: token probabilities supervise generation along student-visited prefixes, while candidate scores and embedding alignment transfer retrieval structure. By initiating generation rollouts from retrieval queries that lack reference answers, ULKD extracts complementary supervision from partially annotated inputs while retaining the original hard objectives. WePM therefore extends prior unified models through coordinated control of visual evidence, heterogeneous supervision, and update magnitude rather than by merely adding another output head.

\vspace{-5pt}
\subsection{Typed Decision and Agentic Control}

Tool-using agents repeatedly select actions from the tools or operations available in their current state. ReAct interleaves language-model reasoning with environment actions~\cite{yao2023react}; Toolformer learns when and how to invoke external APIs from self-supervised signals~\cite{schick2023toolformer}; and Gorilla focuses on selecting and calling from large API collections~\cite{patil2023gorilla}. These approaches broaden language models from response generation to interactive control, but the selected action is typically expressed autoregressively as a tool name, argument sequence, or action string. The surrounding application must then parse and validate that generated output before execution. Typed decision models provide a complementary interface for cases in which the application already defines the valid output space. Jev receives application state together with typed questions and returns probability-bearing Choice, Score, or Noul results rather than free-form text~\cite{typesafejev}. This formulation separates bounded judgment from open-ended response generation: the caller specifies the alternatives or criteria, and the resulting label or score can be consumed directly by program logic. Such decisions can support tool routing, action selection, candidate assessment, and stopping conditions across agentic workflows.

WePM adopts this typed-decision abstraction for on-device use. A schema-conditioned decision slot summarizes the shared multimodal state, and lightweight typed readouts score the caller-defined output space in a single forward pass. Unlike a standalone decision service, this path reuses the same compact vision--language backbone as representation and generation, reducing duplicated resident capacity while preserving a bounded interface for agent control.

\vspace{-5pt}
\subsection{Visual Token Compression and Multi-Granularity Modeling}

High-resolution vision--language models employ several strategies to control the cost of long visual sequences. LLaVA-UHD partitions an image into variable-sized slices and compresses visual tokens while preserving spatial organization~\cite{xu2024llavauhd}, whereas Qwen2.5-VL adopts a native dynamic-resolution vision transformer with window attention~\cite{bai2025qwen25vl}. TokenPacker injects fine-grained local features into compact coarse queries~\cite{li2024tokenpacker}, and VisionZip selects informative tokens before they enter the language model~\cite{yang2024visionzip}. In contrast, inference-oriented approaches such as AdaVSkip reduce computation by bypassing selected transformer modules, without directly reallocating the input-token budget~\cite{sun2026adavskip}.

MGVB is most closely related to input-side visual token compression, but differs in how the retained capacity is organized. It explicitly partitions a fixed budget into global, regional, and fine-grained banks and conditions regional and patch selection on the accompanying text. When text is unavailable, as in offline gallery indexing, MGVB uses a query-independent context to prevent indexed representations from depending on future queries. This design targets heterogeneous evidence requirements across the three capabilities: global semantics support image matching, whereas OCR, detail-centric generation, and visually grounded decision may require localized visual evidence.

\vspace{-5pt}
\subsection{Multitask Optimization under Memory Constraints}

Multitask learning is often affected by gradient interference and unequal optimization rates. PCGrad modifies gradient directions by projecting away conflicting components~\cite{yu2020pcgrad}; FAMO adapts task weights according to relative loss progress~\cite{liu2023famo}; and LDC-MTL balances tasks through loss-discrepancy control~\cite{xiao2025ldcmtl}. These methods address complementary aspects of optimization interference. In compact heterogeneous multitask training, however, the update rule must additionally account for unequal task exposure and the memory cost of retaining heterogeneous computation graphs.

SEMO addresses these practical constraints with fixed per-task coefficients, a predefined exposure schedule derived from the task pools, and sequential gradient accumulation. In our controlled comparison, this design provides a stronger joint trade-off than the tested progress-driven FAMO variants while avoiding their post-update loss re-evaluation. Sequential execution computes the specified joint gradient without retaining all task graphs simultaneously. SEMO is therefore complementary to general gradient-manipulation methods: it targets stable and memory-feasible execution across heterogeneous task pools rather than proposing a universal replacement for adaptive multitask optimization.

\vspace{-5pt}
\section{Limitations and Future Work}
\label{sec:limitations-future-work}

WePM currently exhibits some degradation in agentic capabilities relative to its original Qwen backbone, as the current post-training pipeline primarily focuses on unifying multimodal capabilities without dedicated agent-oriented training. Dedicated post-training to preserve and strengthen tool-use and multi-step interaction capabilities alongside representation, generation, and typed decision has not yet been explored in this work. Multitask optimization also leaves room for improvement: SEMO relies on static analysis of the training-data mixture to determine task weights and exposure schedules, rather than adapting these configurations to evolving learning dynamics. Such dependence may limit its flexibility as task composition and data distributions change.

Future work could explore three directions. First, task-adaptive parameter activation could allow the same model to use different amounts of active capacity according to task complexity and accuracy requirements. Tasks with less stringent requirements could obtain usable results from a subset of parameters, potentially reducing computation and improving response latency. Second, simplifying and extending post-training on top of WePM is important for practical adoption. Lightweight adaptation methods could enable users to incorporate task-specific data and refine targeted capabilities without repeating the full, complex multitask training pipeline or substantially degrading other capabilities. Third, reinforcement-learning methods tailored to unified models could jointly improve heterogeneous capabilities, with particular attention to restoring and strengthening agentic behavior while balancing task quality, cross-task interference, and on-device resource constraints.

\vspace{-5pt}
\section{Conclusion}
\label{sec:conclusion}

We introduce and formalize heterogeneous capability density (HCD) to characterize capability breadth and balanced task quality per unit of resident inference memory, and present WePM, a family of pocket models that unify multimodal representation, generation, and decision within a single vision--language backbone. MGVB reconciles heterogeneous evidence requirements by allocating a fixed visual-token budget across global, regional, and fine-grained banks. ULKD enriches separately collected representation and generation data with soft teacher guidance, avoiding costly cross-task paired annotations and arbitrary hard targets for open-ended inputs. SEMO regulates each task's contribution to shared-parameter updates through explicit loss weights and exposure schedules, while sequential gradient accumulation executes the joint update within training-memory constraints. Together, these mechanisms support normalized embeddings, autoregressive responses, and schema-constrained decisions without duplicating the backbone across capabilities. In the joint representation--generation setting shared with existing unified methods, WePM outperforms all compared baselines across five evaluation groups and raises the Joint score from the strongest baseline's 38.15 to 52.74. In single-task evaluations, WePM-4B achieves the highest average performance across all three capability families, scoring 72.86 over 30 embedding benchmarks, 62.41 over 39 generation benchmarks, and 61.11 over seven typed-decision benchmarks. Compared with the separate-model stack with the highest HCD, WePM achieves slightly higher balanced performance with 61.9\% less resident weight memory, resulting in a 2.64-fold higher HCD (0.244 versus 0.092). These results demonstrate that a compact shared backbone can consolidate diverse multimodal capabilities while maintaining strong and balanced task-specific performance, positioning WePM as a promising foundation for capable on-device AI under tight memory constraints.

\bibliography{paper-refs}

%
%
%
%
%
%
%

\end{document}